\pdfoutput=1

\documentclass[11pt]{article}

\usepackage[preprint]{acl}
\usepackage{times}
\usepackage{latexsym}
\usepackage[T1]{fontenc}

\usepackage[utf8]{inputenc}

\usepackage{microtype}

\usepackage{inconsolata}
\usepackage{graphicx}
\usepackage{tabularx}
\usepackage{multirow}
\usepackage{booktabs}
\usepackage{listings}
\usepackage{xcolor}
\usepackage{pifont}

\newcommand\methodname{\textsc{RingSQL}}
\newcommand\datasetname{\textsc{RingSQL-Gen}} 
\newcommand\datasetnametable{RingSQL-Gen}   

\title{\methodname{}: Schema-Independent Synthetic Data Generation for Text-to-SQL Reinforcement Learning}

\author{
  \textbf{Marko Sterbentz}\textsuperscript{1}
  \quad \textbf{Kevin Cushing}\textsuperscript{2}
  \quad \textbf{Cameron Barrie}\textsuperscript{1}
  \quad \textbf{Kristian J. Hammond}\textsuperscript{1}
  \\
  \textsuperscript{1}Northwestern University
  \quad \textsuperscript{2}Purdue University
  \\
  \texttt{\{marko.sterbentz, cameron.barrie\}@u.northwestern.edu}\\
  \texttt{kcushin@purdue.edu}\\
  \texttt{kristian.hammond@northwestern.edu}
}

\newcommand\satyrn{\textsc{Satyrn}}

\begin{document}
\maketitle

\begin{abstract}
Recent advances in text-to-SQL have been driven by larger models, better datasets, and new training methods like RLVR. However, progress remains limited by scarce high-quality training data, a problem RLVR is especially sensitive to since noisy data can produce spurious rewards. Manual data creation is expensive, and existing synthetic methods trade off reliability for scalability: template-based approaches guarantee correct SQL but need schema-specific templates and lack diversity, while LLM-based generation scales easily but lacks quality guarantees. We introduce \methodname{}, a hybrid framework for generating question-SQL pairs that combines schema-independent query templates with LLM-based paraphrasing of natural language questions. By grounding question generation in complete template questions, \methodname{} preserves question-query correctness across all levels of query complexity, a property purely LLM-based methods fail to maintain. \methodname{} also produces the only synthetic dataset that improves RLVR training performance across all tested model architectures and benchmarks, achieving 69.8\% average accuracy and surpassing both the next-best synthetic dataset by 2.1\% and human-annotated data from Spider and BIRD. Code and data are available at \url{https://github.com/nu-c3lab/RingSQL}.
\end{abstract}

\section{Introduction}

\begin{figure}[!ht] 
\centering 
\includegraphics[width=\columnwidth]{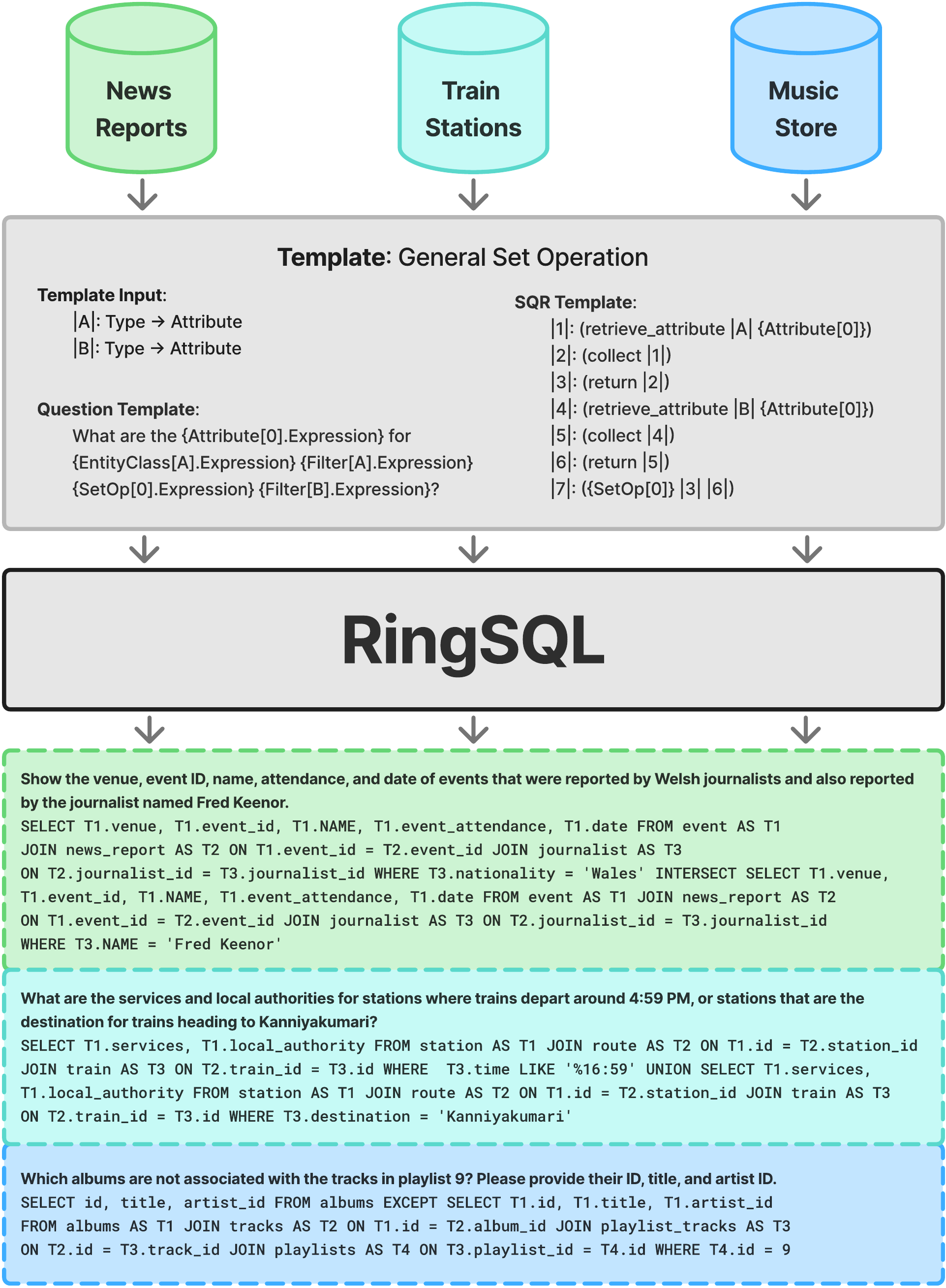}
\caption{\methodname{} applies schema-independent templates for a given question class to any database, generating training questions and queries from its elements.} 
\label{fig:gen_example} 
\end{figure}

\begin{figure*}[ht]
\centering 
\includegraphics[width=\linewidth,keepaspectratio]{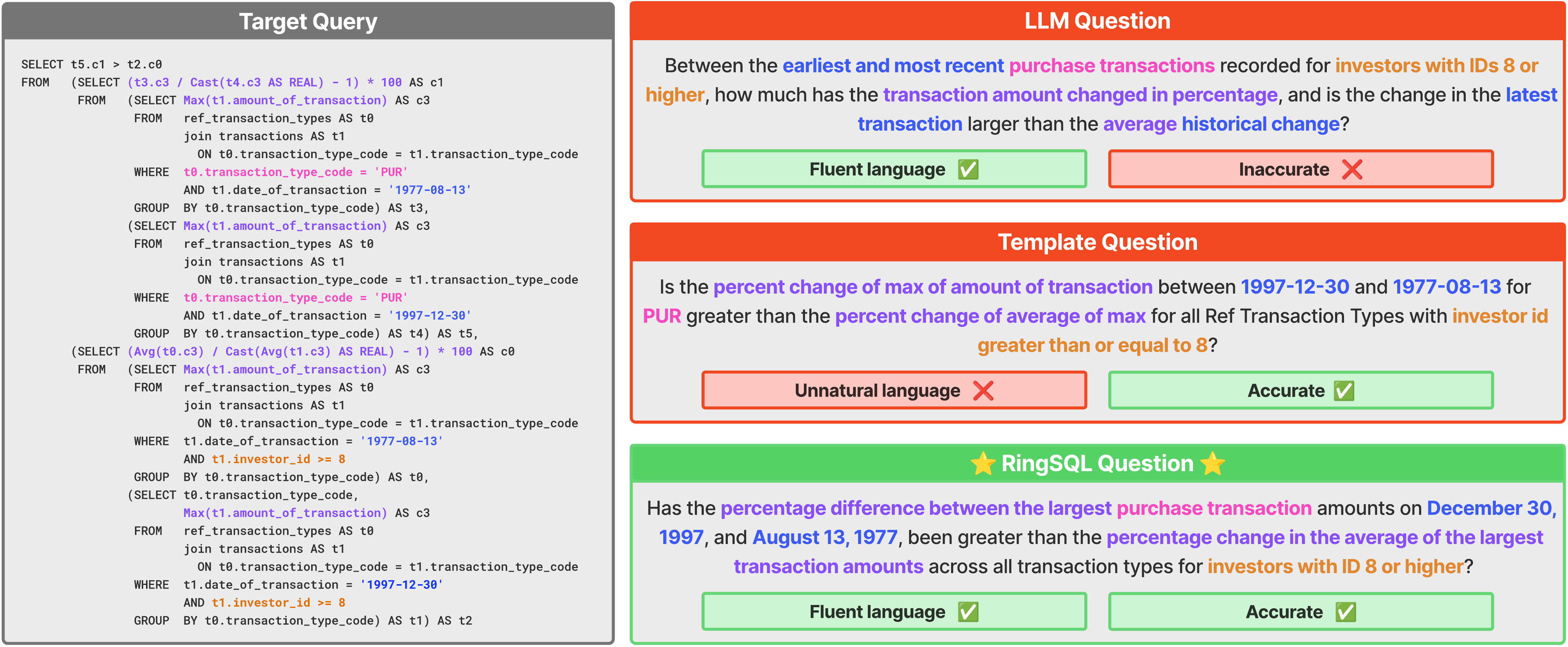}
\caption{Comparison of question generation methods on a complex aggregation query. LLM-only generation is fluent but semantically inaccurate—misrepresenting dates, aggregations, and filters, and hallucinating constraints. Template-based generation is faithful but unnatural. \methodname{} achieves both fluency and faithfulness.} 
\label{fig:motivating_example} 
\end{figure*}

Text-to-SQL systems enable non-experts to interact with databases using natural language, making data access more intuitive and widely available \cite{zelle1996learning, shi2025survey}. These methods have advanced significantly in recent years, driven by new training datasets, increasingly powerful neural models, and new training techniques like reinforcement learning with verifiable rewards (RLVR) \cite{shao2024deepseekmath}.

However, progress in RLVR-based training remains constrained by the limited availability of high-quality data. Recent work shows RLVR is highly sensitive to noise and errors in training data, which can substantially degrade performance \cite{zhu2026noisy}. Even slight mismatches between question and query can result in spurious rewards that harshly penalize correct outputs or promote incorrect outputs. Since human annotation at scale is costly, there is a growing need for synthetic data generation methods that can match or exceed the quality of manually curated datasets.

Two primary paradigms exist for generating synthetic text-to-SQL data: template-based and LLM-based generation. Template- and grammar-based methods construct SQL query templates or abstract syntax trees alongside schema-specific natural language question templates \cite{yu-etal-2018-syntaxsqlnet, yu2020grappa}, offering strong guarantees of syntactic validity and schema compliance. However, designing and maintaining diverse templates across multiple databases is labor-intensive, limiting scalability. Schema-independent templates \cite{weir2020dbpal} reduce this burden but introduce additional post-processing challenges to produce correct joins for a given schema. In all cases, templated language tends to be rigid and unnatural, limiting its utility for training on real-world queries.

LLM-based generation, in contrast, uses large language models to synthesize query-question pairs with minimal human intervention \cite{yang-etal-2024-synthesizing, omnisql2025}, enabling fast, flexible scaling to new domains. However, the lack of oversight introduces risk: generated SQL may be semantically or structurally incorrect, and questions may not accurately reflect query intent. This tradeoff between diversity and reliability is particularly harmful for RLVR-based training \cite{zhu2026revisql}. Since template- and LLM-based methods offer complementary strengths and limitations, combining them offers a promising path forward (Figure \ref{fig:motivating_example}).

In this paper, we present \methodname{}, a hybrid approach that combines the structural reliability of query templates with the scalability and linguistic flexibility of LLMs. \methodname{} uses templates to ensure syntactic and semantic correctness across different databases, while an LLM paraphrases the associated questions to enhance linguistic diversity. This mitigates two major limitations of traditional template-based methods: the need to manually craft templates for each database, and the rigid, repetitive phrasing of templated questions.

With \methodname{}, templates representing specific question types, such as aggregations, joins, or nested conditions, are written once and applied to any database schema, regardless of query complexity (Figure \ref{fig:gen_example}). Rather than building bespoke templates per database, \methodname{} turns each template into a reusable tool for generating diverse, high-quality training data at scale.

Our main contributions are as follows:
\begin{itemize}
    \item We introduce \methodname{}, a template-based approach for generating synthetic SQL queries that generalizes across databases without requiring custom templates for each one, and leverages LLMs to generate natural language questions with rich linguistic variation. Using \methodname{}, we produce \datasetname{}, a text-to-SQL dataset containing 5000 question-query pairs spanning 160 diverse databases.
    \item We show that \datasetname{} achieves higher correctness than LLM-based data generation while maintaining comparable fluency and diversity, properties that are critical for effective RLVR training.
    \item We use \datasetname{} for RLVR training and find a +2.1\% average accuracy improvement on six widely used benchmarks over the next best synthetic dataset, and a +1.3\% improvement over human-annotated data.
\end{itemize}

\section{Related Work}

\paragraph{Synthetic Data Generation for Text-to-SQL}
A wide range of synthetic data generation methods have been proposed to address data scarcity in text-to-SQL training and evaluation. Template- and grammar-based approaches \cite{yu-etal-2018-syntaxsqlnet, weir2020dbpal, yu2020grappa} offer strong control and correctness guarantees, targeting specific query structures while ensuring syntactic validity and schema compliance. However, these methods are labor-intensive, costly to scale, and often produce unnatural language. They are also commonly tailored to individual databases, particularly in domain-specific benchmarks \cite{wang2020text, lee2022ehrsql, kumar-etal-2024-booksql, zhang2024finsql}, requiring manually designed templates per schema and limiting scalability.

Another line of work generates natural language questions from SQL queries produced by some other process \cite{guo-etal-2018-question, zhong-etal-2020-grounded, wang-etal-2021-learning-synthesize, wu-etal-2021-data, guo-etal-2025-sqlforge}. While this can provide weak supervision, there is no guarantee that generated questions faithfully capture query intent, especially for complex queries. More recent methods generate both questions and queries using large language models \cite{yang-etal-2024-synthesizing, omnisql2025, cai2026text2sql}, enabling rapid and diverse data creation with natural language, but without correctness guarantees. In contrast, our approach combines the correctness and controllability of template-based methods with the flexibility and fluency of LLMs, while remaining scalable across new schemas.

\paragraph{Training Text-to-SQL Models}
Beyond advances in dataset construction, text-to-SQL performance has improved markedly due to progress in large language models and training techniques \cite{zhu2024large, shi2025survey}. Early sequence-to-sequence models encode both the question and database schema to generate SQL queries directly \cite{zhong2017seq2sql}. Subsequent work introduced improvements such as constrained decoding \cite{scholak-etal-2021-picard}, relation-aware attention \cite{wang-etal-2020-rat}, schema linking and skeleton parsing \cite{li2023resdsql}, and graph-aware modeling \cite{li2023graphix}. The rise of large autoregressive language models \cite{achiam2023gpt, team2023gemini} has further enabled prompting- and multi-agent-based approaches, including DIN-SQL \cite{pourreza2023din}, DAIL-SQL \cite{gao2024text}, CHESS \cite{chess}, and Chase-SQL \cite{Pourreza2024CHASESQLMR}. More recently, preference- and reinforcement-based training methods such as DPO \cite{rafailov2023direct} and GRPO \cite{shao2024deepseekmath} have been proposed to improve reasoning by encouraging intermediate chain-of-thought generation prior to SQL synthesis \cite{liu-etal-2025-uncovering, ma2025sql, pourreza2025reasoning, yao2025arctic, sterbentz-etal-2026-mixed}. These methods are effective, but rely heavily on data quality and diversity, a central focus of our approach.
\begin{figure*}[ht]
\centering 
\includegraphics[width=\linewidth,keepaspectratio]{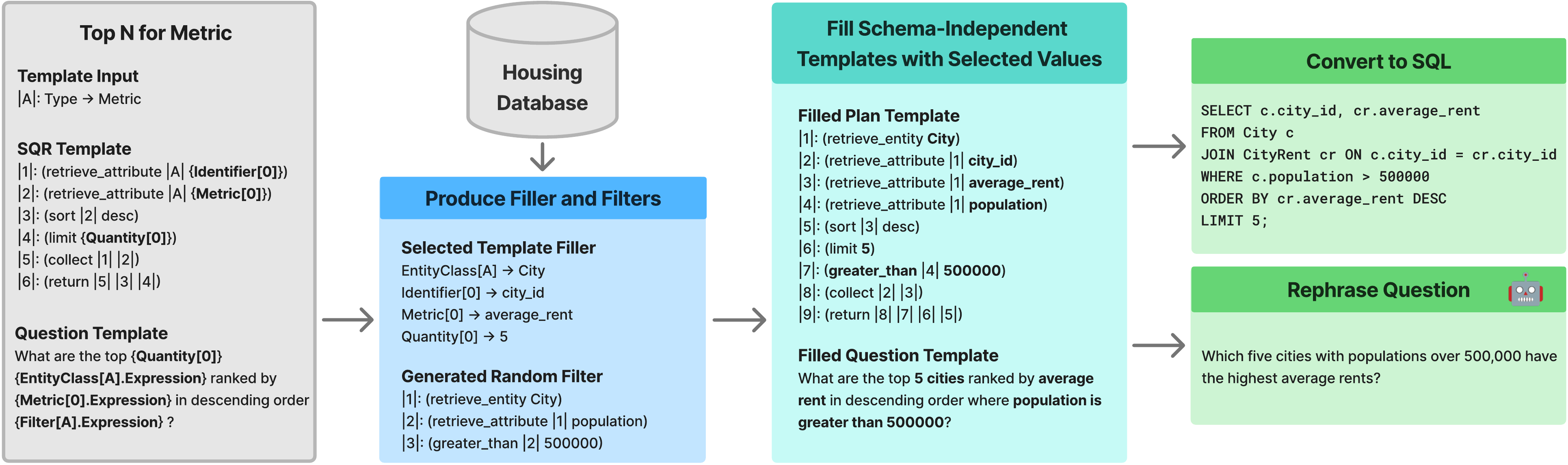}
\caption{The high level approach of \methodname{}. Schema-independent templates are filled in with schema-specific values. Random filters are generated in order to produce a wider variety of questions and incorporated into the filled plan template. The filled plan templates are then deterministically converted to SQL to produce the final query. An LLM is used to produce more natural sounding question based on the filled question template.} 
\label{fig:sagen_process} 
\end{figure*}

\section{Methods}

In this section, we present \methodname{}, our approach for generating high quality text-to-SQL training data on a given relational database. The high level process is shown in Figure \ref{fig:sagen_process}. Schema-independent query templates are at the core of this approach as they provide analytic plans that can be filled with domain specific information before being directly converted to a corresponding SQL query. These query templates each have one or more question templates that are filled in to produce a question that the associated SQL query can answer. The questions are rephrased using an LLM in order to produce more natural sounding language and diverse questions. Randomized filters can be generated and added to the filled query templates in order to generate a more diverse set of queries and questions, as well as to make them better reflect real world querying scenarios. A step-by-step example of how \methodname{} generates a question-query pair can be seen in Appendix \ref{app:query_template_example}.

\subsection{Schema-Independent Templates}

For the query template representation, we adopt Structured Question Representation (SQR) proposed by \cite{sterbentz-etal-2024-satyrn}. Rather than operating directly on the underlying database schema as with SQL, SQR operates on the entities and attributes defined by a Ring. Rings are a simple semantic labeling of the database that captures three key aspects of the data: 1) the entities described by the data, their attributes, and the relationships between them, 2) how to access them in the database based on their corresponding tables and columns, and 3) the natural language names of entities and their attributes. The first is used to fill in the SQR templates, the second in the mapping of SQR templates to specific SQL queries, and the third in the generation of language for questions. Entities correspond to one or more tables, attributes to the columns of these tables, and relationships to the joins between them. The simple nature of the Rings allows them to be created programmatically based on the database DDL, as we do in our experiments.

The use of Rings and SQR templates, rather than SQL templates, decouples data access, which is specific to the database schema, from data analysis. This formulation allows for schema-specific information, such as joins, to be abstracted away when defining query templates. This makes it ideal for defining templates that are reusable across diverse database schemas.

\subsubsection{Template Structure}

Query templates are defined by three components: the input, the SQR template, and one or more question templates. A fully specified example query template can be seen in Figure \ref{fig:query-template-appendix} of Appendix \ref{app:query_template_example}.

The inputs to the query templates are one or more attributes of an entity in a Ring. These define the primary columns and data to be used in the query. Inputs can have mandatory filters defined for them which is useful for specifying if an individual or cohort with a specific property should be retrieved in the data and later used during analysis in the SQR template.

The SQR template is a complete SQR plan with various components replaced by slots to be filled in. These slots are filled by one or more random values including attributes of a Ring entity, random values derived from the database, and sorting direction strings. Each slot can have additional labels that help indicate what value should fill it. For example, \textit{Expression} indicates the slot should be filled with the natural language description of an attribute, and  \textit{Value} indicates the slot should be filled with a specific value of an attribute. Each slot in a SQR template is uniquely identifiable in the plans. 

These schema-independent templates make it easy to target specific question types. However, training robust text-to-SQL models also requires question–query pairs that span a wide range of query structures. To address this, we complement the fixed templates with a set of generators that support variable numbers of inputs and relationships. These generators are also schema-independent, but they dynamically instantiate SQR template structures by varying the number of input attributes or by enforcing constraints on how many relationships are used to select those attributes. Depending on the types of relationships used, these generators also apply groupings and aggregations to attributes to produce a reasonable query. This approach yields question–query pairs that cover a broader set of query structures. These generators fill slots, add filters, and generate questions in the same way as the regular SQR templates.

\subsubsection{Filling the SQR Plan Templates}

SQR template inputs are filled by randomly selecting a set of attributes that satisfy the type constraints on each input. For example, if an input is constrained to an \textit{Arithmetic} type, it is filled with an attribute that is represented as a numeric value. This ensures all subsequent operations performed on each input in the SQR plan are plausible. 

To ensure that all slots are filled with valid values and no conflicts arise (e.g. using an attribute for an entity that has not been defined yet), the values to fill each slot are chosen in a particular order. First, the entities to use are selected, then attributes of these entities are selected to use as input, and finally any specific values required from those attributes are sampled from the database. This allows \methodname{} to ensure reasonable values are used for comparison by examining the actual values for a column and their distribution to ensure that comparison and filter values make sense within the context of the data. The slot filling process proceeds in this way until a complete SQR plan has been produced.

\subsubsection{Filter Generation}

Once a SQR template is filled, filters can be added to increase the complexity and diversity of the data produced by \methodname{}. Filters are generated in one of two ways. The first is by randomly sampling an attribute, a filter operation, and a value to create a simple filter (e.g. \texttt{WHERE <column> = <value>}). This is done by first sampling an attribute from a given entity and an appropriate operation based on the type of the sampled attribute. A value for the attribute is then sampled from the database if needed. The original SQR plan is then augmented with this filter, and the generated questions are updated to include a natural language phrase expressing the filter. These simple filters can be combined with \texttt{AND} and \texttt{OR} operations to increase complexity.

For the second method of generating filters, we create SQR filter templates which can be filled in a similar manner to SQR plan templates (e.g. \texttt{WHERE <column> > (SELECT avg(<column>) FROM <table> WHERE <simple-filter>)}). In many cases, the input for these filters is fixed by the SQR plan's input assignments, although sometimes sampling is used to fill these. These templates can also be combined with a simple filter to produce an even more complex filter, as shown in the example. This template-based approach makes it easy to define new complex filters along with more concise natural language expression templates for them.

\subsection{Schema-Specific SQL Generation}

\paragraph{Converting SQR to SQL} Once we have the executable SQR plans, we need to convert them to SQL. We first compile the SQR plans to remove any duplicate retrieval steps. We then utilize the analysis engine of \satyrn{} \cite{sterbentz-etal-2024-satyrn} which executes SQR plans by converting them to SQL queries against the underlying database. We pass each SQR plan generated by \methodname{} through this analysis engine to produce the SQL queries for our dataset.

\subsection{Question Generation}

To complete the text-to-SQL dataset, we also generate a natural language question for each synthesized SQL query. Questions are initially produced using question templates paired with each query template: when input slots in a SQR template are instantiated, the corresponding slots in the associated question template are filled to yield a natural language question.

While this templated approach ensures semantic alignment between questions and SQL queries, the resulting language may not reflect how a human would phrase it and is often not grammatical. While additional language templates can partially mitigate this issue, template-based generation still tends to produce rigid or unnatural phrasing. To address this, we apply a lightweight rephrasing step using an LLM. We first use the LLM to generate a brief description of each database based on the schema and first three rows of every table, which helps disambiguate poorly named or ambiguous columns. For each generated question, we then prompt the LLM to rephrase the template-based question given the original question, its corresponding SQL query, the database description, and three similar question–query examples from Spider. This encourages more natural, domain-specific phrasing that better reflects how a human might ask the question. The full rephrasing prompt is shown in Figure \ref{fig:full_rephrasing_prompt} of Appendix \ref{app:question_generation_ablations}.

\section{Experimental Setup}
\paragraph{Training Datasets} We study the effect of RLVR when training on data generated by \methodname{}. To construct this data, we manually design 32 query templates and generators spanning a range of complexity, from simple retrieval and aggregation to ranking and cross-cohort comparisons over time.

We generate Rings for the Spider train and dev databases using a script that assumes a one-to-one mapping between tables and entities, columns and attributes, and relationships and joins, with joins automatically inferred from foreign keys. Given each Ring–database pair, \methodname{} produces question–SQL pairs. The result is \textbf{\datasetname{}}, a dataset of 5,000 question–SQL pairs (4,000 train, 1,000 dev, following the Spider database splits), with questions rephrased using GPT-4o-mini.

We compare against three synthetic data generation methods: \textbf{SENSE} \cite{yang-etal-2024-synthesizing}, SynSQL \cite{omnisql2025}, and \textbf{SQLFlow} \cite{cai2026text2sql}, sampling 5,000 examples from each. For SynSQL, we use two subsets: \textbf{SynSQL-Uniform}, sampled uniformly across difficulty levels, and \textbf{SynSQL-Complex}, the same 5,000 complex question–SQL pairs used by \cite{ma2025sql} to train state-of-the-art models. We also compare against models trained on samples of the Spider and BIRD train sets to assess synthetic data against human-annotated data. For all datasets, each SQL query is executed to verify a non-null result, ensuring a meaningful reward signal during RLVR training.

\paragraph{Evaluation Datasets}
We evaluate on two widely used benchmarks, \textbf{Spider} \cite{spider} and \textbf{BIRD} \cite{bird}, to test our method's efficacy in training state-of-the-art models. We additionally evaluate on three Spider variants: \textbf{Spider-DK} \cite{gan-etal-2021-exploring}, \textbf{Spider-Syn} \cite{gan-etal-2021-towards}, and \textbf{Spider-Realistic} \cite{deng-etal-2021-structure}, which test robustness to linguistic variation, questions with fewer exact table/column mentions, and greater reliance on domain knowledge.

\paragraph{Base Models and Training}
We train models of varying sizes, both coding and non-coding: Qwen2.5-Coder-3B-Instruct, Qwen2.5-Coder-7B-Instruct \cite{hui2024qwen25codertechnicalreport}, Llama-3.2-3B-Instruct, and Llama-3.1-8B-Instruct \cite{grattafiori2024llama3herdmodels}. We use RLVR with the reward function from \cite{ma2025sql}, which specifies progressive rewards for output format, SQL syntax, query correctness, and response brevity. Training uses 4 A100 80GB GPUs for 1000 steps with a batch size of 4. Our primary metric is execution accuracy. Each model generates 8 samples at temperature 0.8, with self-consistency used to select the final SQL.
\begin{table*}[ht!]
\centering
\footnotesize
\setlength{\tabcolsep}{5pt}
\begin{tabular}{llccccccc}
\toprule
\textbf{Base Model} & \textbf{RL Data} 
  & \shortstack{\textbf{Spider}\\\textbf{Dev}} 
  & \shortstack{\textbf{Spider}\\\textbf{Test}} 
  & \shortstack{\textbf{BIRD}\\\textbf{Dev}} 
  & \shortstack{\textbf{Spider}\\\textbf{DK}} 
  & \shortstack{\textbf{Spider}\\\textbf{Syn}} 
  & \shortstack{\textbf{Spider}\\\textbf{Realistic}} 
  & \textbf{Avg.} \\
\midrule

\multirow{8}{*}{\shortstack[l]{Llama-3.2\\3B-Instruct}}
& None                & 64.22 & 66.33 & 28.29 & 55.51 & 53.87 & 54.92 & 53.86 \\
\cmidrule(lr){2-9}
& SynSQL-Unif         & 59.19 & 62.55 & 24.84 & 49.16 & 49.61 & 50.00 & 49.23 \\
& SynSQL-Comp         & 53.97 & 55.89 & 27.77 & 48.79 & 46.81 & 47.64 & 46.81 \\
& SENSE               & 56.96 & 62.93 & 30.25 & 48.41 & 52.61 & 53.54 & 50.78 \\
& SQL-Flow            & 61.22 & 66.60 & 28.49 & 54.39 & 54.45 & 54.72 & 53.31 \\
& \datasetnametable{} & \textbf{65.57} & \textbf{69.68} & \textbf{33.12} & \textbf{61.31} & \textbf{57.64} & \textbf{57.87} & \textbf{57.53} \\
\cmidrule(lr){2-9}
& Spider              & \underline{68.57} & \underline{70.14} & 30.50 & 57.20 & \underline{60.64} & \underline{60.83} & \underline{57.98} \\
& BIRD                & 63.15 & 67.68 & 32.07 & 56.64 & 57.16 & 57.68 & 55.73 \\
\midrule

\multirow{8}{*}{\shortstack[l]{Qwen2.5-Coder\\3B-Instruct}}
& None                & 77.85 & 79.60 & 47.46 & 66.73 & 66.05 & 70.08 & 67.96 \\
\cmidrule(lr){2-9}
& SynSQL-Unif         & 78.82 & 78.99 & \textbf{53.13} & 68.04 & 69.34 & 72.83 & 70.19 \\
& SynSQL-Comp         & 80.46 & 78.71 & 52.67 & 69.72 & 69.05 & 71.65 & 70.38 \\
& SENSE               & 78.72 & 80.76 & 46.22 & 68.79 & 68.38 & 71.65 & 69.09 \\
& SQL-Flow            & 80.75 & 82.30 & 52.86 & 71.03 & 72.15 & \textbf{73.82} & 72.15 \\
& \datasetnametable{} & \textbf{82.69} & \textbf{82.53} & \textbf{53.13} & \textbf{71.96} & \textbf{72.24} & 73.62 & \textbf{72.70} \\
\cmidrule(lr){2-9}
& Spider              & 77.27 & 79.74 & 46.74 & 68.04 & 68.67 & 69.09 & 68.26 \\
& BIRD                & 79.30 & 80.48 & 51.62 & 69.35 & 70.31 & \underline{73.82} & 70.81 \\
\midrule

\multirow{8}{*}{\shortstack[l]{Llama-3.1\\8B-Instruct}}
& None                & \underline{78.05} & 78.30 & 49.67 & \underline{70.09} & 69.54 & \underline{75.00} & 70.11 \\
\cmidrule(lr){2-9}
& SynSQL-Unif         & 72.92 & 76.15 & 46.67 & 67.10 & 68.47 & 71.06 & 67.06 \\
& SynSQL-Comp         & 76.02 & 77.88 & 50.98 & 67.29 & 68.09 & 73.23 & 68.92 \\
& SENSE               & 76.60 & 76.06 & 47.32 & 60.93 & 67.31 & 72.24 & 66.74 \\
& SQL-Flow            & 77.18 & 77.22 & 48.56 & 68.04 & 69.44 & \textbf{75.00} & 69.24 \\
& \datasetnametable{} & \textbf{77.47} & \textbf{79.04} & \textbf{53.00} & \textbf{70.09} & \textbf{70.79} & 73.23 & \textbf{70.60} \\
\cmidrule(lr){2-9}
& Spider              & 77.18 & 78.48 & 48.76 & 68.60 & 69.44 & 72.64 & 69.18 \\
& BIRD                & 76.40 & 78.25 & 52.48 & 69.16 & 68.86 & 73.82 & 69.83 \\
\midrule

\multirow{8}{*}{\shortstack[l]{Qwen2.5-Coder\\7B-Instruct}}
& None                & 85.11 & 86.26 & 56.26 & 73.64 & 77.56 & 81.89 & 76.79 \\
\cmidrule(lr){2-9}
& SynSQL-Unif         & 82.01 & 83.23 & 58.87 & 72.34 & 73.02 & 77.56 & 74.51 \\
& SynSQL-Comp         & 84.43 & 85.51 & 61.15 & \textbf{76.45} & 76.40 & \textbf{82.87} & 77.80 \\
& SENSE               & 78.05 & 80.07 & 53.91 & 67.85 & 72.15 & 73.43 & 70.91 \\
& SQL-Flow            & 83.95 & 83.47 & 60.36 & 73.64 & 75.15 & 78.94 & 75.92 \\
& \datasetnametable{} & \textbf{85.50} & \textbf{86.77} & \textbf{61.73} & 75.70 & \textbf{78.72} & 81.89 & \textbf{78.39} \\
\cmidrule(lr){2-9}
& Spider              & 85.40 & 86.40 & 60.62 & 76.07 & 78.43 & \underline{83.46} & \underline{78.40} \\
& BIRD                & 85.11 & 86.68 & 57.82 & \underline{76.82} & \underline{78.82} & 80.71 & 77.66 \\

\bottomrule
\end{tabular}
\caption{Execution accuracy (\%) on Spider dev/test, BIRD dev, and three Spider robustness variants (DK, Syn, Realistic). Within each model, rows are grouped as: no RL training (top), synthetic RL data (middle), and gold-standard RL data (bottom, upper bound). Best result among synthetic data per model shown in \textbf{bold}. Results for non-synthetic data that perform at least as well as all synthetic data tuned models are \underline{underlined}.}
\label{tab:primary_results}
\end{table*}

\section{Results}

\subsection{\datasetname{} Delivers Consistent Gains Across Models and Benchmarks}

\datasetname{} achieves the highest average accuracy for every model architecture among synthetic datasets, and is the \textit{only} one to improve performance for every model tested, with gains ranging from 0.49--4.74\%. Full results of training on each dataset and evaluating across six text-to-SQL benchmarks appear in Table \ref{tab:primary_results}. 

\begin{figure*}[ht]
\centering 
\includegraphics[width=\linewidth,keepaspectratio]{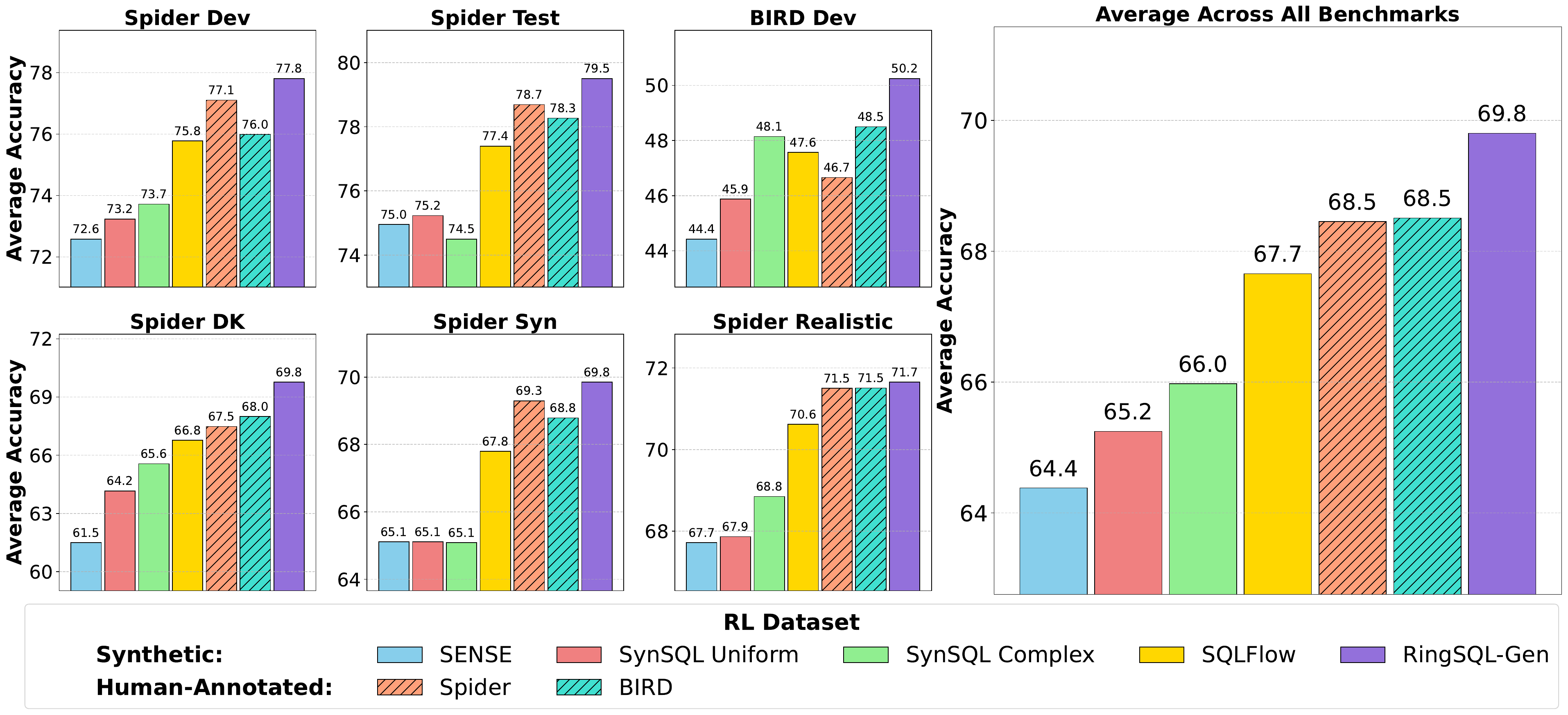}
\caption{Average accuracy (\%) across six text-to-SQL benchmarks for models trained on each RL dataset, aggregated over all model architectures. Human-annotated datasets are shown with hatched bars. \datasetname{} achieves the highest accuracy across all benchmarks, outperforming other synthetic and human-annotated datasets.} 
\label{fig:avg_accuracy_all} 
\end{figure*}

Aggregating across all architectures and benchmarks (Figure \ref{fig:avg_accuracy_all}), \datasetname{} reaches 69.80\% average accuracy, outperforming every other synthetic dataset by margins of 2.1\% (SQLFlow, the next best) to 5.4\% (SENSE, the weakest). Against the base models' overall average of 67.18\%, only \datasetname{} and SQLFlow improve performance, with gains of +2.62\% and +0.48\%, respectively; the remaining three (SENSE, SynSQL-Uniform, SynSQL-Complex) degrade overall capability on the Llama models while improving it on Qwen2.5-Coder-3B, and results are mixed on Qwen2.5-Coder-7B, where only \datasetname{} and SynSQL-Complex improve overall capability.

Together, these results show \datasetname{} produces a more broadly useful training signal than existing synthetic alternatives, delivering consistent gains across model sizes and benchmark domains.

\subsection{\datasetname{} Outperforms Human-Annotated Training Data}
To contextualize \methodname{}'s synthetic data quality, we separately train on 5,000 human-annotated examples from Spider and and 5,000 sampled from BIRD, comparing each against training on \datasetname{}'s 5,000 examples. Full results can be seen in Table \ref{tab:primary_results}. Training on Spider and BIRD improves average performance by +1.27\% and +1.33\% over the untrained baseline (67.18\%), both trailing \datasetname{}'s +2.62\% gain. \datasetname{} is the only synthetic dataset to surpass \emph{either} human-annotated dataset in overall average accuracy: 69.80\% for \datasetname{} vs. 68.45\% for Spider and 68.51\% for BIRD. 

Human-annotated data retains some in-distribution advantage. Spider-trained models excel on Spider-derived benchmarks, and BIRD-trained models beat most synthetic datasets on BIRD Dev. However, \datasetname{} achieves the best overall average without such specialization, indicating a training signal that is both higher quality in aggregate and more broadly transferable.

\subsection{\methodname{} Balances Question Correctness, Fluency, and Diversity}

We evaluate LLM-generated questions along three axes: correctness and fluency using UniEval \cite{zhong-etal-2022-towards}, and linguistic diversity using two syntactic metrics (\% unique POS sequences, \% unique dependency graphs) and two lexical metrics (MATTR \cite{covington2010cutting}, MTLD \cite{mccarthy2010mtld}). Correctness is measured via consistency and relevance against template-generated questions as reference; fluency via naturalness and understandability. Tables \ref{tab:correctness_fluency_results} and \ref{tab:linguistic_diversity_results} summarize these results across all datasets.

Grounding the LLM in template-based questions, as \datasetname{} does, substantially improves correctness: consistency increases from 54.45 to 75.07, and relevance from 52.30 to 74.07, compared to a zero-shot LLM given only the SQL query and schema. This confirms that our schema-independent templates help ensure generated questions remain faithful to the underlying SQL, and this correctness advantage holds across query complexity, while zero-shot LLM performance degrades as complexity increases (see Appendix \ref{app:complexity} for more details). This gain in correctness comes with only a modest fluency cost: all LLM-generated datasets achieve similarly high fluency scores (89--91), except SENSE ($\sim$78) and the human-annotated Spider and BIRD datasets, which score lower since annotators prioritized query coverage over natural phrasing. Purely template-based questions score lowest (52--54) on fluency.

Table \ref{tab:linguistic_diversity_results} shows that this correctness-fluency balance does not come at the expense of diversity. Templated questions consistently score lowest across all linguistic diversity metrics, while \datasetname{} has comparable diversity with the best purely LLM-generated datasets, showing that our hybrid framework overcomes the rigidity typical of template-based approaches. SENSE and Spider, both rigidly phrased for direct question-query alignment, show the lowest diversity overall (extended discussion in Appendix \ref{app:linguistic_diversity}). Query diversity statistics, which can be found in Table 5 of Appendix \ref{app:training_data_stats}, show a similar pattern: \datasetname{} and SynSQL-Complex show comparable diversity. SynSQL-Uniform is more diverse, thanks to its broader mix of simple and complex queries, while SQLFlow is slightly less diverse.

\begin{table}[ht!]
\centering
\scriptsize
\setlength{\tabcolsep}{5pt}
\begin{tabular}{lcccc}
\toprule
 & \multicolumn{2}{c}{\textbf{Correctness}} & \multicolumn{2}{c}{\textbf{Fluency}} \\
\cmidrule(lr){2-3} \cmidrule(lr){4-5}
\textbf{RL Data} & \textbf{Consist.} & \textbf{Relev.} & \textbf{Natural.} & \textbf{Underst.} \\
\midrule
Templ. Questions Only & \multicolumn{2}{c}{\textit{reference}} & 53.61 & 52.39 \\
\midrule
LLM 0-Shot w/o Templ.            & 54.45 & 52.30 & \textbf{91.75} & \textbf{91.36} \\
LLM 0-Shot w/ Templ.  & 73.21 & 71.86 & 91.38 & 90.99 \\
\datasetnametable{}   & \textbf{75.07} & \textbf{74.07} & 89.31 & 88.74 \\
\midrule
SynSQL-Uniform        & \multicolumn{2}{c}{---} & 89.44 & 89.63 \\
SynSQL-Complex        & \multicolumn{2}{c}{---} & 89.28 & 89.61 \\
SENSE                 & \multicolumn{2}{c}{---} & 78.99 & 77.53 \\
SQLFlow               & \multicolumn{2}{c}{---} & 91.34 & 90.92 \\ 
\midrule
Spider              & \multicolumn{2}{c}{---} & 78.95 & 77.48 \\
BIRD               & \multicolumn{2}{c}{---} & 83.83 & 82.85 \\
\bottomrule
\end{tabular}
\caption{Correctness (consistency, relevance) and fluency (naturalness, understandability) for each RL training dataset. Higher is better for all measures.}
\label{tab:correctness_fluency_results}
\end{table}

\begin{table}[ht!]
\centering
\scriptsize
\setlength{\tabcolsep}{2pt}
\begin{tabular}{lcccc}
\toprule
 & \multicolumn{2}{c}{\textbf{Syntactic Diversity}} & \multicolumn{2}{c}{\textbf{Lexical Diversity}} \\
\cmidrule(lr){2-3} \cmidrule(lr){4-5}
\textbf{RL Data} & \textbf{\% Uniq. POS} & \textbf{\% Uniq. Dep.} & \textbf{MATTR} & \textbf{MTLD} \\
\midrule
Templ. Questions Only & 81.03 & 94.73 & 0.6650 & 35.62 \\
LLM 0-Shot w/o Templ.            & 94.25 & 99.27 & 0.7880 & 75.21 \\
LLM 0-Shot w Templ.  & \textbf{94.67} & 99.22 & 0.7848 & 72.11 \\
\midrule
\datasetnametable{}   & 92.79 & 98.89 & 0.7661 & 63.45 \\
SynSQL-Uniform        & 93.00 & \textbf{99.75} & 0.7633 & 64.24 \\
SynSQL-Complex        & 89.77 & 99.60 & 0.7556 & 60.71 \\
SENSE                 & 54.64 & 86.37 & 0.7355 & 51.91 \\
SQLFlow               & 89.30 & 99.00 & \textbf{0.8125} & \textbf{90.89} \\
\midrule
Spider               & 53.84 & 85.99 & 0.7385 & 52.25 \\
BIRD              & 69.15 & 95.50 & 0.7728 & 70.98 \\
\bottomrule
\end{tabular}
\caption{Syntactic and lexical diversity metrics for each RL training set. Higher scores indicate greater diversity.}
\label{tab:linguistic_diversity_results}
\end{table}

\subsection{Additional Results}
Question generation ablations (App. \ref{app:question_generation_ablations}) show that rephrasing templated questions, using them as a reference for final question generation, and including few-shot examples in the generation prompt all improve text-to-SQL capabilities. We also find that augmenting existing datasets with \methodname{} data yields stronger models (App. \ref{app:augmentation_study}). We further provide studies on SQR template diversity (App. \ref{app:sqr_diversity}) and an evaluation on Spider 2.0-lite (App. \ref{app:spider_2}).
\section{Conclusion}

In this work, we present \methodname{}, an approach for generating synthetic SQL query-question pairs using schema-independent templates that are reusable across database schemas. By grounding LLM paraphrasing in template-generated questions, \methodname{} achieves the correctness of template-based methods alongside the fluency and diversity of LLM-based generation. Using \methodname{}, we construct \datasetname{}, a 5,000-pair dataset spanning 160 databases, and models trained with RLVR on \datasetname{} outperform those trained on prior synthetic and human-annotated data across six benchmarks. These results suggest that combining the structural guarantees of templates with the flexibility of LLMs offers a promising path toward scalable, high-quality data generation for RLVR-based training.

\section*{Ethical Considerations}
This work focuses on generating synthetic training data for text-to-SQL reasoning models trained via RLVR. All examples are fully synthetic and are constructed using query templates grounded in specific database schemas. Our experiments use databases from the publicly available Spider benchmark, which are released under permissive licenses and, to our knowledge, do not contain personal, proprietary, or sensitive information. Large language models are used only to rephrase synthetic natural language questions and do not operate on real user queries or introduce personally identifiable information.

Improving text-to-SQL capabilities may lower the barrier to querying structured databases, which could be misused in downstream systems if deployed without appropriate access controls or auditing. Additionally, synthetic data generation may reflect biases in the underlying templates, benchmark schemas, or rephrasing models, potentially underrepresenting certain domains or interaction styles. These risks primarily arise from deployment and data selection choices rather than the generation method itself.

\section*{Limitations}

While \methodname{} can be used to generate queries for a variety of SQL dialects, we generate and utilize only SQLite queries as this is the type of database used in the text-to-SQL benchmarks we used for evaluation. Our work focuses on data quality for RL-based training paradigms like RLVR. We conducted preliminary experiments applying the same synthetic data to supervised fine-tuning and observed inconsistent results across datasets; a full investigation of SFT-specific factors (data scale, hyperparameters, output formatting) is left to future work. Additionally, due to computational constraints, we only trained models ranging from 3B to 8B parameters. It would be worth investigating the usage of our training data for improving the capabilities of larger models.

\section*{Acknowledgments}

We would like to thank the Center for Advancing the Safety of Machine Intelligence (CASMI) for funding this work. This research was supported in part through the computational resources and staff contributions provided for the Quest high performance computing facility at Northwestern University which is jointly supported by the Office of the Provost, the Office for Research, and Northwestern University Information Technology. We also thank the ACL ARR reviewers for their valuable comments and constructive feedback.

\bibliography{anthology,custom}

\appendix

\section{Data Generation with a Query Template}
\label{app:query_template_example}

\subsection{Plan Templates}
\begin{figure}[!tbp] 
\centering 
\includegraphics[width=\columnwidth]{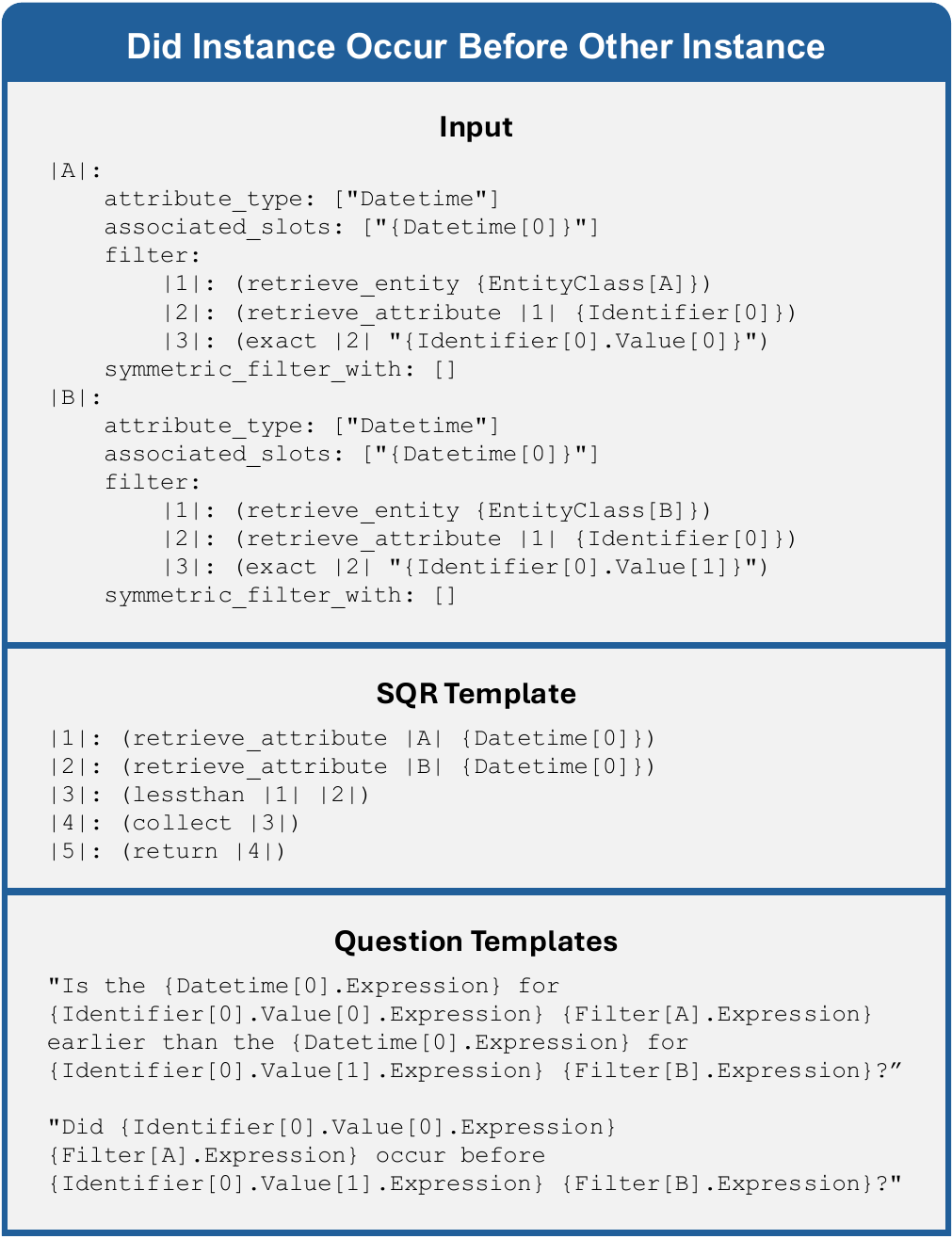}
\caption{An example of the schema-independent query template for generating queries that check if one instance occurred before another.}
\label{fig:query-template-appendix} 
\end{figure}

An example of the components that make up a SQR plan template is shown in Figure \ref{fig:query-template-appendix}. The plan input defines initial subplans that return attributes to be used in the main SQR template. These subplans include slots with specified types to be filled by sampling available attributes for an entity. The example contains only one such slot \texttt{\{Datetime[0]\}} which can be filled by \texttt{Datetime} type attributes. Including indexes in defined slots provides a convenient way to define multiple distinct slots of the same type which can be filled with different attributes. In the example, since both plan inputs use the same slot, they will be filled with the same sampled attribute. The filters for each input specify the required filter operations that should be applied to the sampled attribute(s) before being passed to the analysis template. The example specifies a filter that returns the sampled attribute(s) for only one row based on its unique identifier value. The \texttt{.Value} suffix indicates that a slot should be filled with a value sampled from the database, and the \texttt{[0]} and \texttt{[1]} indices in this example indicate that these values should be filled separately. The optional value for \texttt{symmetric\_filter\_with} provides a way to specify that the same filters should be used for multiple plan inputs. The SQR template defines the analysis to be performed on the results from the plan inputs. In this example, the analysis is simply a comparison between the two \texttt{Datetime} values returned by the plan inputs. The question templates provide multiple possible phrasings of a templated question that the plan template answers. The \texttt{.Expression} suffix denotes that the slot should be filled with the name of the attribute or entity, the natural language description of the generated filter, or a string representation of the sampled value.

\begin{figure*}[htbp]
\centering 
\includegraphics[width=\linewidth,keepaspectratio]{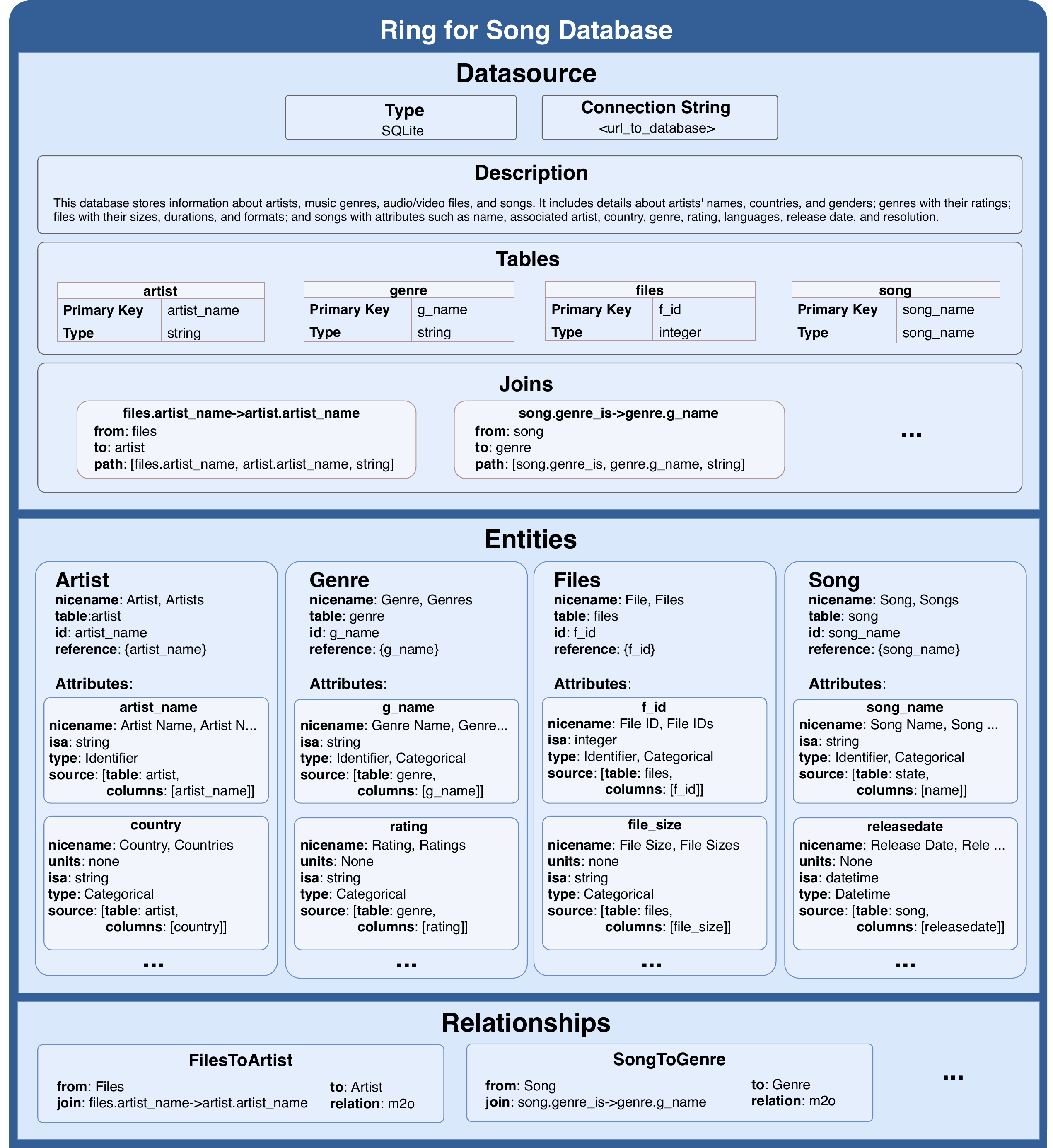}
\caption{An example of a Ring for a song database. This Ring contains four entities along with their attributes. These are used fill the slots of the SQR template during the data generation process.} 
\label{fig:ring_diagram} 
\end{figure*}

\begin{figure}[!tb] 
\centering 
\includegraphics[width=\columnwidth]{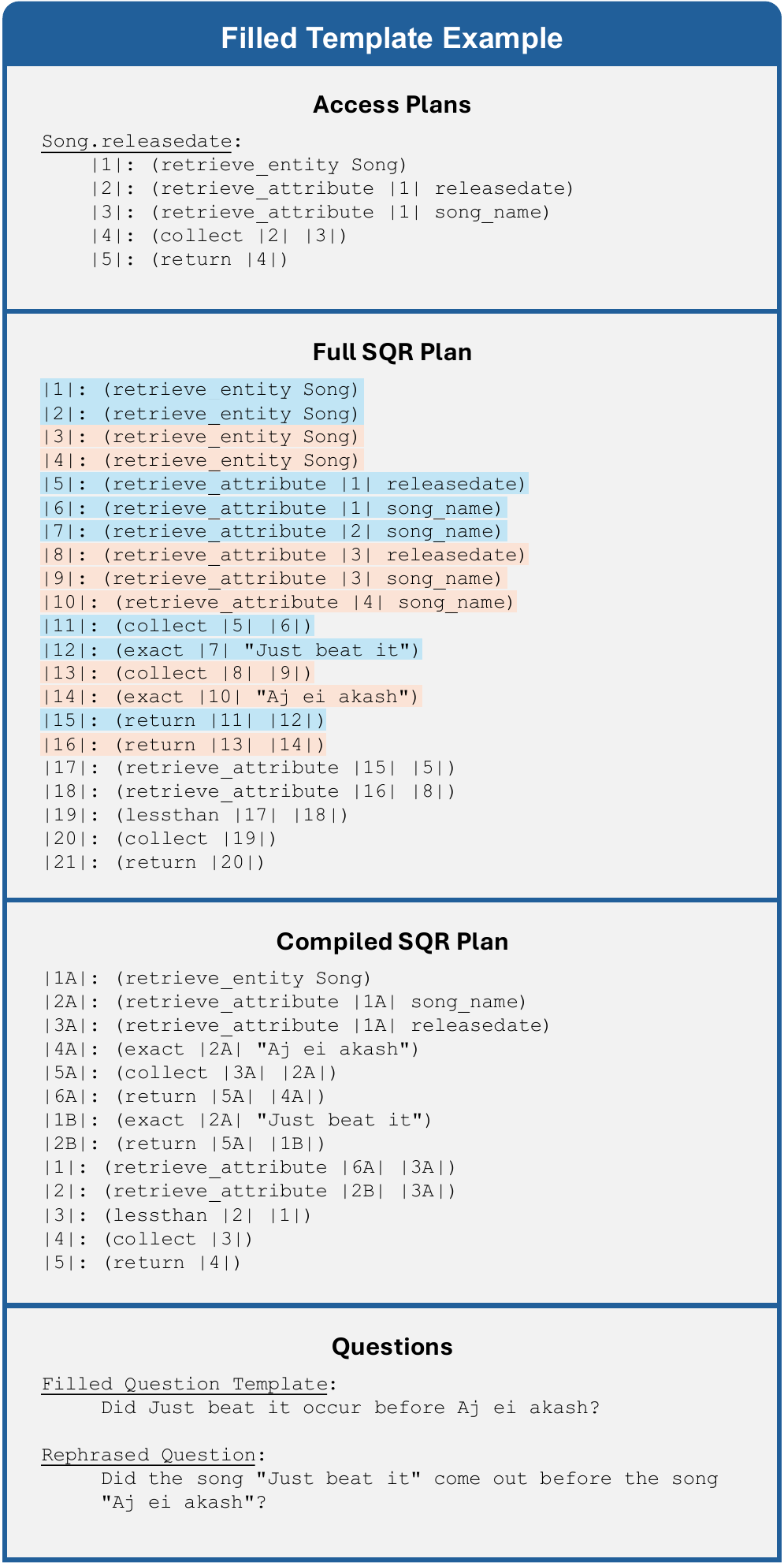}
\caption{An example of how the schema-independent template from Figure \ref{fig:query-template-appendix} is filled in to produce an intermediate SQR plan and corresponding question. The blue and orange highlighted steps denote the steps generated for the plan inputs \texttt{|A|} and \texttt{|B|} respectively.}
\label{fig:filled-template-appendix} 
\end{figure}

\begin{figure*}[htbp] 
\centering 
\includegraphics[width=\linewidth,keepaspectratio]{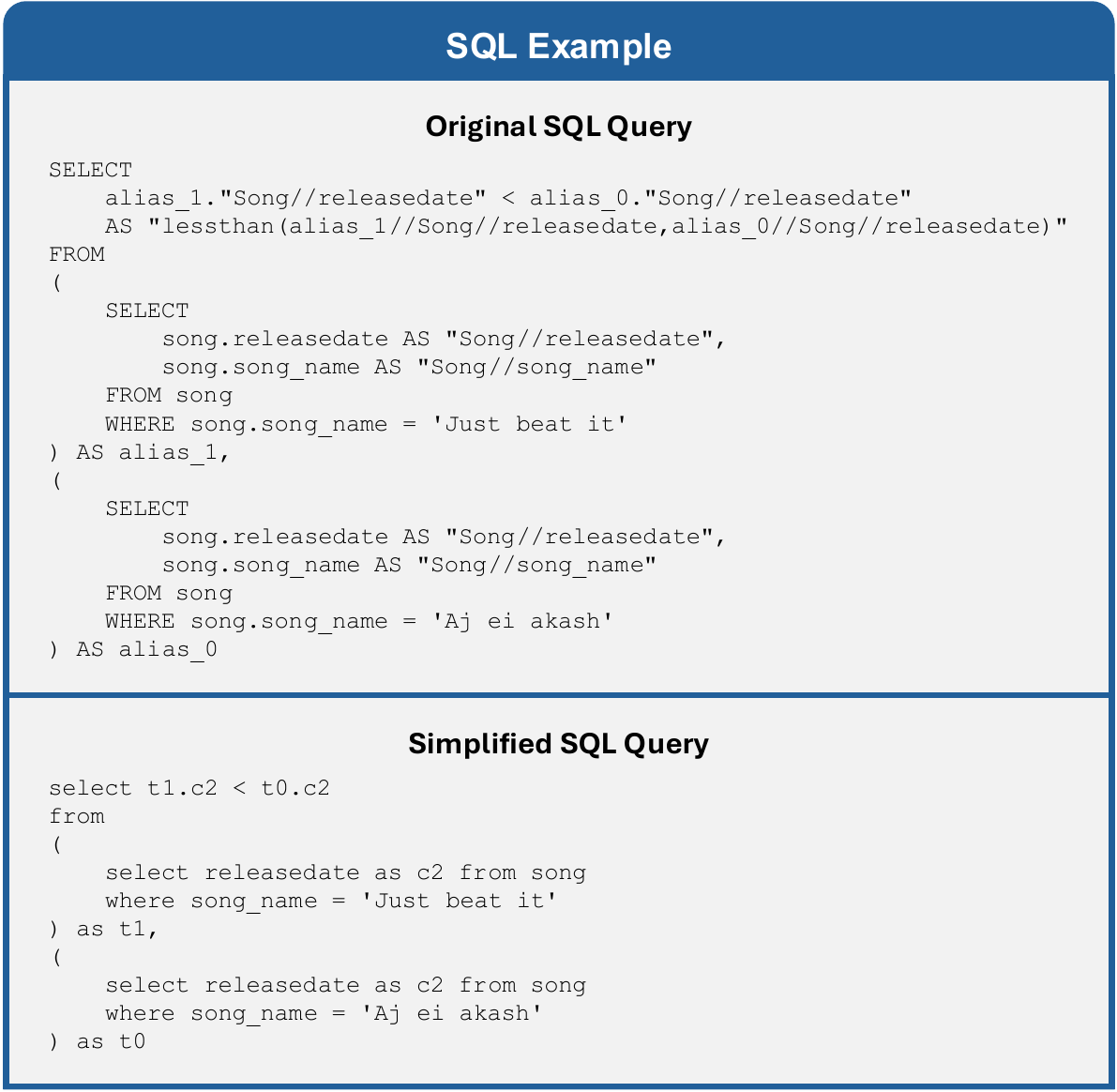}
\caption{An example of the initial SQL query and its simplified form based on the compiled SQR plan in Figure \ref{fig:filled-template-appendix}.}
\label{fig:final_sql_example} 
\end{figure*}

\subsection{Filling the Plan Template}

The slots specified in the plan inputs are filled first by selecting an entity and attribute from a Ring \cite{sterbentz-etal-2024-satyrn, sterbentz2023lightweight, paley2021data}. In this example, the Ring and the available entities and attributes are shown in Figure \ref{fig:ring_diagram}. Suppose the \texttt{releasedate} attribute of the \texttt{Song} entity is selected to fill the \texttt{\{Datetime[0]\}} slot. Other attributes, like \texttt{song\_name} would not be valid options for this slot since it is a categorical attribute and it is not valid to compare whether one category occurred prior to another in this template. The ability to impose these type constraints on sampled attributes ensures that the queries generated by \methodname{} only contain reasonable operations and correctly answer the filled template questions. The access plan shown in Figure \ref{fig:filled-template-appendix} gives the SQR plan to retrieve the attribute assigned to the slot. This assignment also assigns the entity class slots for both input filters to the \texttt{Song} entity. Next, the filter slots are filled in, first the \texttt{\{Identifier[0]\}} slot is filled with the identifying attribute of the \texttt{Song} entity (the \texttt{song\_name} attribute), then the \texttt{\{Identifier[0].Value\}} slots are filled in with values sampled from the \texttt{song\_name} attribute (\texttt{"Just beat it"} and \texttt{"Aj ei akash"}). After all slots are filled, additional filters can be generated and appended to plan inputs to increase the complexity of the query. Then, the access plans, plan inputs, and analysis template are combined into the full SQR plan shown, and the slots in the question templates are filled based on the slot assignments. The resulting plan has many redundant steps, which are cleaned up by the SQR compiler, producing the compiled SQR plan shown. A single filled template question for the plan is sampled to be used for rephrasing. As shown in the example, the template question is correct with respect to the plan but is not reflective of how a person would ask the question, while the rephrased question retains the meaning and sounds much more natural.

\subsection{Simplifying the SQL Query}
Once the initial queries are produced, we pass each of them through a process to simplify the query as much as possible, while still retaining the original execution output of the query. First, the query is parsed into an abstract syntax tree. The simplifier traverses this tree and performs a variety of simplifications, including removing unused column and table aliases, stripping table prefixes from single table subqueries, providing joined tables with simpler aliases, and inlining subqueries when possible. In this process, we also standardize queries to be of the form \texttt{SELECT \_ FROM \_ WHERE \_ ...} and convert queries to lower case with the exception of inline string values. Since the original queries are composed sequentially from the SQR plan steps, they often contain redundancies and complex aliases that are not reader-friendly. These modifications result in queries that have a more readable standard structure and are $\sim$25\% shorter than the original queries on average. 

After the SQR plan previously described is compiled, it is converted into the SQL query in Figure \ref{fig:final_sql_example}. While the initial query executes successfully, it includes long aliases, an unnecessary alias in the outer \texttt{SELECT} statement, and unnecessary columns in the sub-queries. The result of our AST-based simplification method on the original query is shown in Figure \ref{fig:final_sql_example}. It is significantly shorter without the unnecessary columns and aliases, making it more readable without changing its result when executed.

\section{Question Generation Ablations}
\label{app:question_generation_ablations}

\begin{figure*}
\centering 
\includegraphics[width=\linewidth,keepaspectratio]{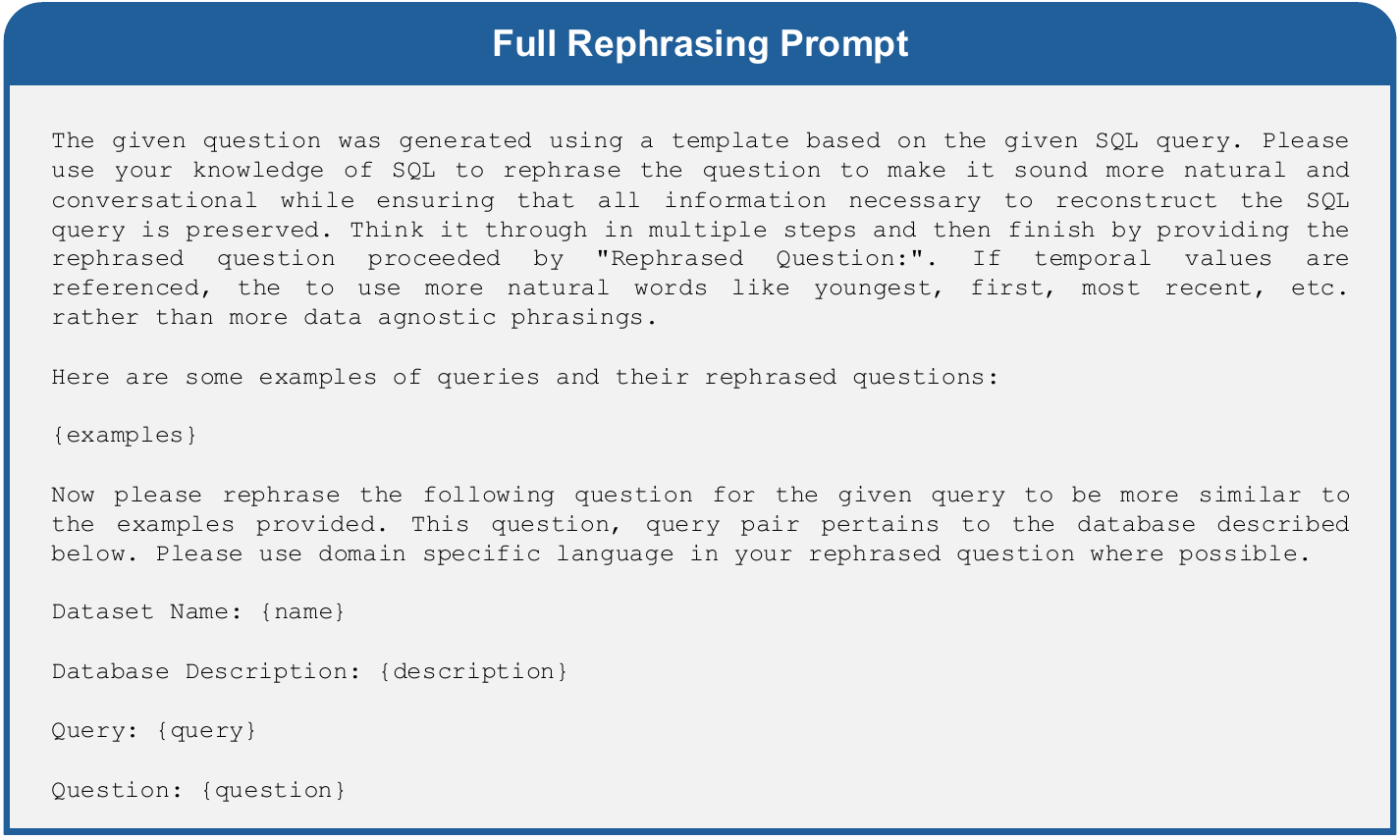}
\caption{The prompt used to rephrase the template-based questions generated by \methodname{}.} 
\label{fig:full_rephrasing_prompt} 
\end{figure*}

\begin{figure*}
\centering 
\includegraphics[width=\linewidth,keepaspectratio]{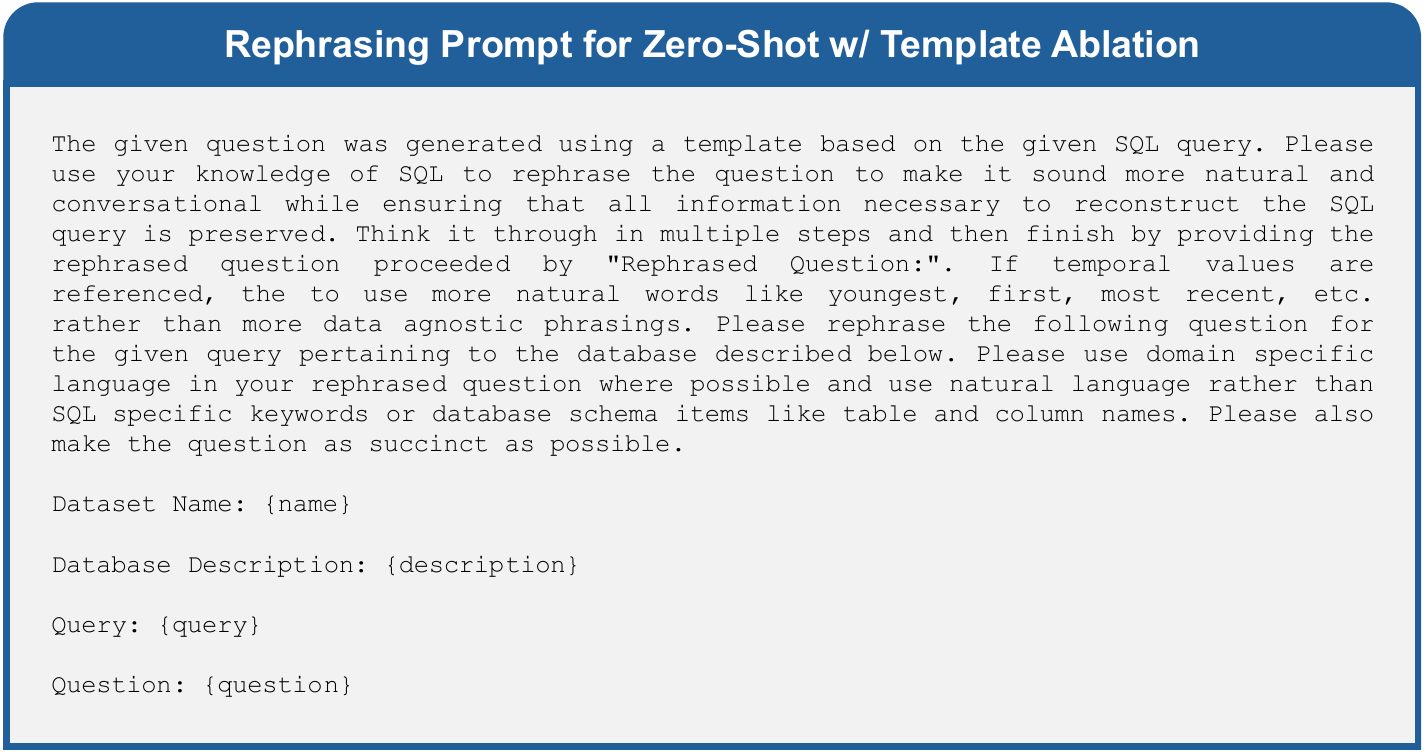}
\caption{The prompt used to rephrase the template-based questions generated by \methodname{}, but without using any question and query examples for in-context learning.} 
\label{fig:no_few_shot_examples_rephrasing_prompt} 
\end{figure*}

\begin{figure*}
\centering 
\includegraphics[width=\linewidth,keepaspectratio]{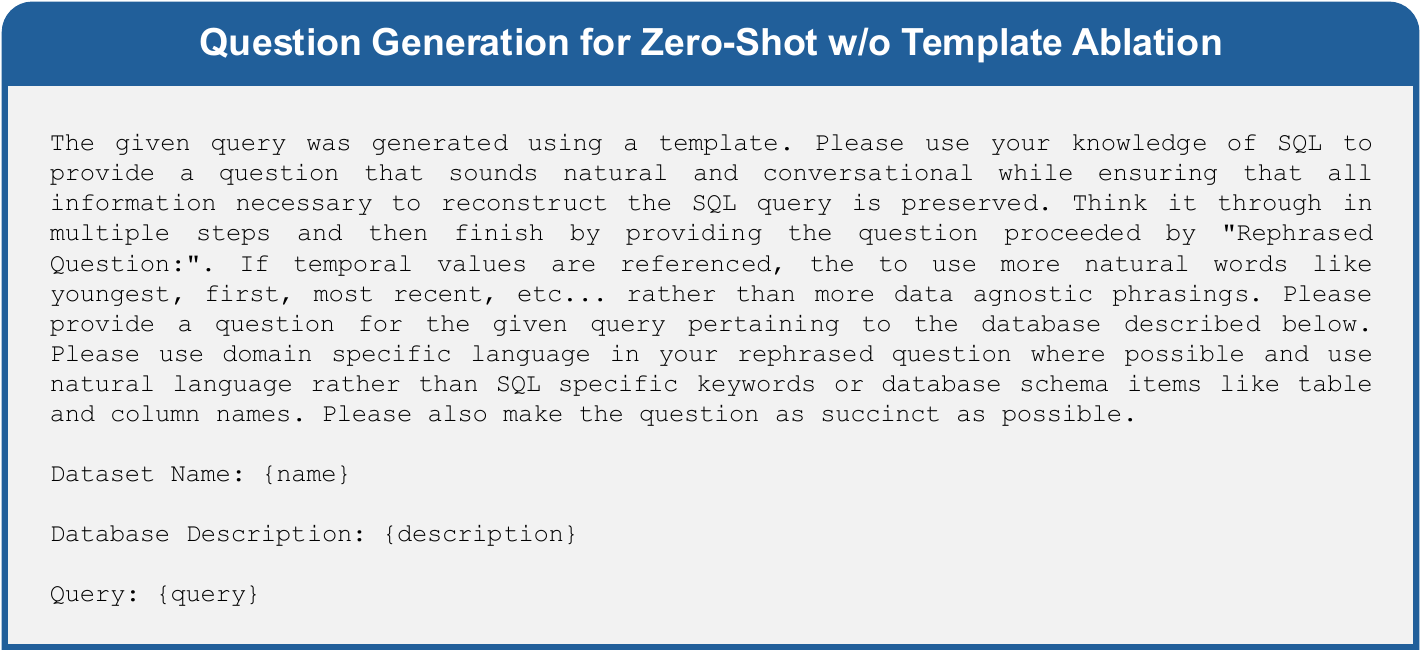}
\caption{The prompt used to generate questions using no template-based questions or few-shot examples.} 
\label{fig:query_only_prompt} 
\end{figure*}

To assess the impact of \methodname{}'s question generation process and language on text-to-SQL capabilities, we perform three ablations. For all ablations, we use the same \datasetname{} queries, only changing the language of the questions for each setting. The prompt used for rephrasing the template-based questions produced with \methodname{} can be seen in Figure \ref{fig:full_rephrasing_prompt}, along with the prompt used in the LLM Zero-Shot w/ Template ablation in Figure \ref{fig:no_few_shot_examples_rephrasing_prompt} and the LLM Zero-Shot w/o Template ablation in Figure \ref{fig:query_only_prompt}. Table \ref{tab:ablation_results} contains the results of training Qwen2.5-Coder-3B-Instruct for each ablated setting.

\begin{table*}[ht!]
\centering
\footnotesize
\begin{tabular}{
    p{0.25\linewidth}
    p{0.07\linewidth}
    p{0.07\linewidth}
    p{0.07\linewidth}
    p{0.07\linewidth}
    p{0.07\linewidth}
    p{0.09\linewidth}
    p{0.09\linewidth}
}
\toprule
\multirow{2}{\linewidth}{\centering\textbf{Setting}} &
\centering\textbf{Spider Dev} &
\centering\textbf{Spider Test} &
\centering\textbf{BIRD Dev} &
\centering\textbf{Spider DK} &
\centering\textbf{Spider Syn} &
\centering\textbf{Spider Realistic} &
\multirow{2}{\linewidth}{\centering\textbf{Average}} \tabularnewline
\midrule
\datasetnametable{}-3B & \centering\textbf{82.69} & \centering\textbf{82.53} & \centering\textbf{53.13} & \centering\textbf{71.96} & \centering{72.24}  & \centering{73.62} & \centering\textbf{72.70} \tabularnewline
\midrule

Templ. Questions Only & \centering{78.34} & \centering{80.07} & \centering{52.87} & \centering{68.97} & \centering{68.47}  & \centering{71.26}  & \centering{69.99} \tabularnewline

LLM Zero-Shot w/o Template & \centering{80.66} & \centering{81.23} & \centering{47.39} & \centering{69.53} & \centering\textbf{73.11}  & \centering\textbf{75.39} & \centering{71.22} \tabularnewline

LLM Zero-Shot w/ Template & \centering{80.46} & \centering{80.48} & \centering{51.43} & \centering{71.59} & \centering{72.53}  & \centering{74.61} & \centering{71.85} \tabularnewline
\bottomrule
\end{tabular}
\caption{Comparison of execution accuracy (\%) on Spider, BIRD, and the Spider variants for each question generation setting. Base model used for training and evaluation in all settings was Qwen2.5-Coder-3B-Instruct.}
\label{tab:ablation_results}
\end{table*}

\subsection{Rephrasing Template-Based Questions Improves Model Performance}

We investigate the impact of \methodname{}'s rephrasing module by removing it entirely and training only on template-based questions generated by directly filling question templates with values. Results of training with these raw template-based questions can be seen in Table \ref{tab:ablation_results} as \textbf{Template Questions Only}. In this setting, accuracy drops consistently across all six benchmarks (69.99\% vs. 72.70\%). This suggests that while template-based questions provide correct semantic supervision, their limited linguistic variability encourages models to overfit to rigid surface forms. Rephrasing introduces more natural and diverse language realizations of the same underlying queries, which better matches the distribution of real user questions and improves the model’s ability to generalize beyond the specific phrasings of the templates. Rephrasing therefore acts as a form of linguistic data augmentation, decoupling semantic understanding from specific lexical patterns and reducing sensitivity to template-specific wording.

\subsection{Template Grounding Improves Question Generation over LLM Alone}

We investigate the impact of grounding the question generation with the template-based questions. In this setting, we directly generate a question from the query instead of rephrasing the template-based question. We refer to this setting as \textbf{LLM Zero-Shot w/o Template} in Table \ref{tab:ablation_results}. The prompt used for this can be seen in Figure \ref{fig:query_only_prompt}. This method of generating synthetic training data is common for LLM-based methods since there are no template-based questions with which to ground the question generation. The average performance of the model trained with these questions across the six benchmarks is 71.22\%, which is lower than the 72.7\% for the \datasetname{} data. Notably, performance of the LLM Zero-Shot w/o Template model is higher on the Spider-Syn and Spider-Realistic benchmarks, but substantially lower on the BIRD Dev set where it performs -5.74\% worse. This suggests that, while direct generation from queries can produce reasonable surface-level questions for simpler queries, it struggles to fully capture the meaning of complex queries. Grounding in template-based questions provides a reliable intermediate representation that ensures all query components are expressed correctly. This template-based grounding acts as a scaffold that preserves the full semantic content of queries and it allows the model to learn the correct mappings more consistently, resulting in a +5.48\% improvement on BIRD compared to LLM Zero-Shot w/o Template.

\subsection{Few-Shot Examples Improve Question Rephrasing}

We investigate the impact of using few-shot examples in the rephrasing prompt by using only a basic rephraser without these examples. The results can be seen as \textbf{LLM Zero-Shot w/ Template} in Table \ref{tab:ablation_results}. The prompt used for this can be seen in Figure \ref{fig:no_few_shot_examples_rephrasing_prompt}. In this setting, average performance is again lower at 71.85\%. However, as with the LLM Zero Shot w/o Template setting, we observe slightly higher performance on the Spider-Syn and Spider-Realistic benchmarks. 
This suggests that even a basic rephraser can introduce linguistic diversity while remaining grounded in the template-generated questions. However, performance decreases on the Spider Dev and Test sets, which contain more questions with more explicit mentions of columns. This indicates that few-shot examples guide the rephraser to produce questions that better align with the phrasing patterns of the benchmarks. In other words, few-shot examples act as stylistic anchors, ensuring that the rephrased questions both preserve the query semantics and mirror the surface-level patterns of target datasets, which improves generalization on the benchmarks. For real-world scenarios, the few-shot examples could be replaced with examples tailored for the target database and ways in which a user would likely ask questions in order to improve the utility of the text-to-SQL interface in practice.


\section{Correctness vs. Query Complexity}
\label{app:complexity}

\begin{figure}[!ht] 
\centering 
\includegraphics[width=\columnwidth]{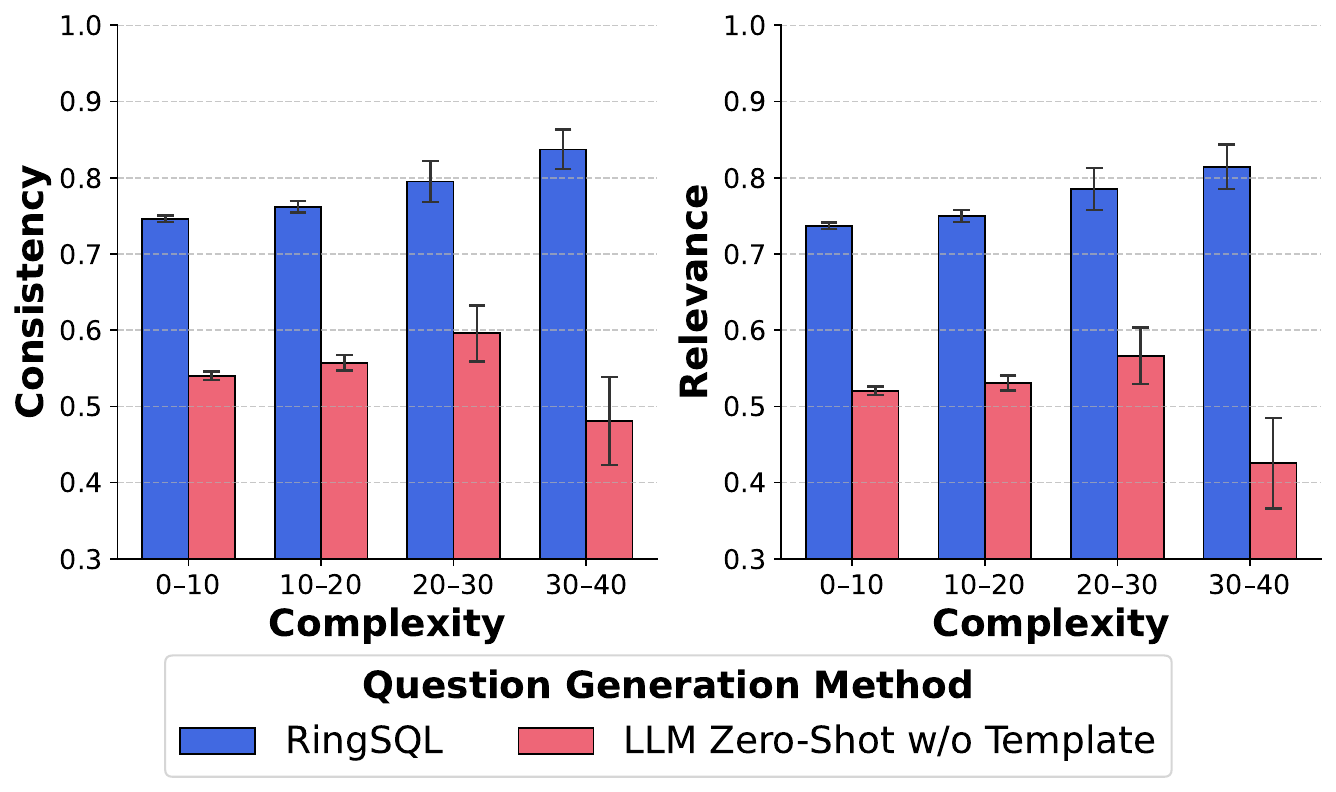}
\caption{Consistency and relevance scores by query complexity for two synthetic question generation methods: RingSQL and LLM Zero-Shot w/o Template. Error bars denote standard error. RingSQL outperforms the pure LLM method across all complexity levels, with the gap widening at higher complexity.} 
\label{fig:correctness_complexity_comparison} 
\end{figure}

We measure query complexity as a weighted sum of joins, filters, aggregations, sorts, and subqueries (weights: 3 for joins, 2.5 for subqueries, 1 otherwise), bucketed in intervals of 10. Figure \ref{fig:correctness_complexity_comparison} compares consistency and relevance across complexity buckets for \methodname{} versus a zero-shot LLM using only the SQL query and schema. Consistency measures the factual alignment between the target question and the reference question. Relevance measures the whether the target question contains only the important information of the reference question.

\methodname{} remains stable across all complexity levels (0.74–0.84), while the zero-shot LLM degrades noticeably as complexity increases, dropping from \textasciitilde0.54 to \textasciitilde0.49 on consistency and \textasciitilde0.52 to \textasciitilde0.43 on relevance by the 30–40 bucket. Notably, the zero-shot LLM already trails by \textasciitilde0.20 points even in the simplest bucket (0–10), indicating that the correctness gap is not purely a function of complexity overload: there is a baseline information loss when templates aren't provided, and the template signal remains valuable across all complexity levels.

\section{Linguistic Diversity Details}
\label{app:linguistic_diversity}

Table \ref{tab:linguistic_diversity_results} in the main text reports full syntactic and lexical diversity metrics for each RL training dataset. Here we discuss these results in more depth.

Across all four metrics, templated questions yield the lowest diversity, scoring significantly below the LLM-generated question sets on \% unique POS sequences, MATTR, and MTLD. RingSQL-Gen exhibits diversity scores similar to both SynSQL splits and SQLFlow, indicating that our hybrid framework achieves parity with purely LLM-based methods, overcoming a central limitation of purely template-based approaches. The SENSE dataset has the lowest syntactic and lexical diversity, similar to Spider, which, while human-annotated, is rigidly phrased to ensure direct alignment between question and query content.

\section{Training Data Properties}
\label{app:training_data_stats}

We report a set of statistics describing the databases, queries, and natural language questions in the training data used in our experiments. Database- and query-level properties are summarized in Table \ref{tab:dataset_properties}. We include the total number of databases, along with the average number of tables and columns referenced per query, which provide an overview of schema usage across the dataset. We also include the number of unique templates the queries represent. This is determined by the inclusion and ordering (including nesting) of structural SQL operations (select, from, where, group by, etc.). We also report the average depth of each query’s abstract syntax tree (AST) as a measure of structural complexity. In addition, we measure the average number of operations per query, as well as the average number of joins and \texttt{WHERE} conditions, which characterize the prevalence of different SQL constructs in the data.

On average, \datasetname{} is more complex with more tables, joins, \texttt{WHERE} conditions, and other SQL operations than the SENSE, SQLFlow, and SynSQL datasets. SynSQL-Complex, composed of only "Complex" queries, is slightly less complex in terms of the average metrics than SynSQL-Uniform, which includes data from all 4 SynSQL difficulty levels. This is caused by some "Highly Complex" queries being included in this sample, indicated by higher maximum values for the number of tables, joins, and where conditions. Although \datasetname{} seems to have less complex outliers, its higher average complexity suggests that it generally consists of more complex queries than the SynSQL datasets. The SQL simplification process we use also causes reductions in metrics like the number of unique templates and AST depth of \datasetname{}. Our experimental results also indicate that these complex outliers do not necessarily contribute to improvements in model performance.

\begin{table*}[ht!]
\small
\centering
\begin{tabular}{
    p{0.15\linewidth}
    p{0.04\linewidth}
    p{0.08\linewidth}
    p{0.08\linewidth}
    p{0.08\linewidth}
    p{0.08\linewidth}
    p{0.08\linewidth}
    p{0.08\linewidth}
    p{0.08\linewidth}
}
\toprule
\multirow{3}{\linewidth}{\centering\textbf{Training Dataset}} &
\multirow{3}{\linewidth}{\centering\textbf{\# DBs}} &
\multirow{3}{\linewidth}{\centering\textbf{\# Tables per SQL}} &
\multirow{3}{\linewidth}{\centering\textbf{\# Cols per SQL}} &
\multirow{3}{\linewidth}{\centering\textbf{\# Unique Templ.}} &
\centering\textbf{AST Depth\\per SQL} &
\multirow{3}{\linewidth}{\centering\textbf{\# Ops\\per SQL}} &
\multirow{3}{\linewidth}{\centering\textbf{\# Joins\\per SQL}} &
\multirow{3}{\linewidth}{\centering\textbf{\# Where per SQL}} \tabularnewline
\midrule

\multirow{2}{\linewidth}{\centering{SENSE}} & \multirow{2}{\linewidth}{\centering{140}} & \centering{1.67\\(1, 10)} & \centering{3.51\\(0, 20)} & \multirow{2}{\linewidth}{\centering{211}}  & \centering{0.07\\(0, 2)}  & \centering{2.47\\(0, 18)}  & \centering{0.54\\(0, 8)} & \centering{1.44\\(1, 5)} \tabularnewline
\midrule
\multirow{2}{\linewidth}{\centering{SynSQL-Uniform}} & \multirow{2}{\linewidth}{\centering{3448}} & \centering{2.17\\(1, 14)} & \centering{7.70\\(0, 57)} & \multirow{2}{\linewidth}{\centering{755}}  & \centering{0.08\\(0, 3)}  & \centering{4.33\\(0, 27)}  & \centering{1.15\\(0, 13)} & \centering{1.83\\(1, 17)} \tabularnewline
\midrule
\multirow{2}{\linewidth}{\centering{SynSQL-Complex}} & \multirow{2}{\linewidth}{\centering{3407}} & \centering{1.98\\(1, 11)} & \centering{6.93\\(0, 71)} & \multirow{2}{\linewidth}{\centering{548}}  & \centering{0.05\\(0, 2)}  & \centering{3.39\\(0, 31)}  & \centering{0.96\\(0, 10)} & \centering{1.51\\(1, 15)} \tabularnewline
\midrule
\multirow{2}{\linewidth}{\centering{SQLFlow}} & \multirow{2}{\linewidth}{\centering{146}} & \centering{1.62\\(1, 9)} & \centering{4.16\\(0, 29)} & \multirow{2}{\linewidth}{\centering{479}}  & \centering{0.18\\(0, 2)}  & \centering{2.58\\(0, 27)}  & \centering{0.47\\(0, 7)} & \centering{1.33\\(1, 7)} \tabularnewline
\midrule
\multirow{2}{\linewidth}{\centering{\datasetnametable{}}} & \multirow{2}{\linewidth}{\centering{139}} & \centering{2.44\\(1, 12)} & \centering{7.33\\(0, 28)} & \multirow{2}{\linewidth}{\centering{570}}  & \centering{0.45\\(0, 2)}  & \centering{5.02\\(0, 22)}  & \centering{1.28\\(0, 10)} & \centering{1.86\\(1, 5)}  \tabularnewline
\midrule

\multirow{2}{\linewidth}{\centering{Spider}} & \multirow{2}{\linewidth}{\centering{140}} & \centering{1.66\\(1, 10)} & \centering{3.49\\(0, 20)} & \multirow{2}{\linewidth}{\centering{209}}  & \centering{0.08\\(0, 2)}  & \centering{2.45\\(0, 18)}  & \centering{0.54\\(0, 8)} & \centering{1.43\\(1, 5)} \tabularnewline
\midrule

\multirow{2}{\linewidth}{\centering{BIRD}} & \multirow{2}{\linewidth}{\centering{68}} & \centering{2.04\\(1, 9)} & \centering{5.01\\(1, 21)} & \multirow{2}{\linewidth}{\centering{473}}  & \centering{0.07\\(0, 2)}  & \centering{4.43\\(0, 22)}  & \centering{1.01\\(0, 6)} & \centering{2.07\\(1, 11)} \tabularnewline

\bottomrule
\end{tabular}
\caption{Properties of the databases and queries of the RL training datasets used in our experiments. The average (min, max) are reported, where applicable.}
\label{tab:dataset_properties}
\end{table*}

\section{Augmenting Human-Annotated Data with Synthetic Data}
\label{app:augmentation_study}

To assess whether data generated with RingSQL can complement existing human-annotated datasets, we conduct a targeted augmentation study in which 1,000 randomly sampled RingSQL-Gen examples are added to each of the Spider and BIRD training samples described above. Given that these experiments are intended to probe whether augmentation provides consistent gains rather than to identify the best-performing model, we run them on Qwen2.5-Coder-3B-Instruct, a mid-tier architecture that is neither the strongest nor weakest model in our evaluation suite and thus well-suited to reflect representative trends. Results can be seen in Table \ref{tab:augmentation_results}.

\paragraph{Spider + RingSQL-Gen} Augmenting the Spider sample with 1,000 RingSQL-Gen examples improves average performance by 3.5 percentage points, from 68.26\% to 71.78\%. Accuracy increases across all benchmark datasets, with a particularly strong gain of 6.7\% on Spider Realistic.

\paragraph{BIRD + RingSQL-Gen} Augmenting the BIRD sample with RingSQL-Gen data produces a slight average decrease of 0.37\%, from 70.81\% to 70.44\%. The decline is largely attributable to a 3.15\% drop on Spider Realistic, while other benchmarks see modest gains.

The asymmetry between the two augmentation settings suggests that data produced by RingSQL can be a productive source of augmentation, as demonstrated by the substantial gains over the Spider baseline, but does not guarantee improvements in all cases. The performance drop observed with BIRD augmentation, concentrated on a single benchmark, points to potential distribution mismatch rather than a general incompatibility, and warrants further investigation into selection strategies for augmentation samples.

\begin{table*}[ht!]
\centering
\begin{tabular}{
    p{0.25\linewidth}
    p{0.07\linewidth}
    p{0.07\linewidth}
    p{0.07\linewidth}
    p{0.07\linewidth}
    p{0.07\linewidth}
    p{0.09\linewidth}
    p{0.09\linewidth}
}
\toprule
\multirow{2}{\linewidth}{\centering\textbf{RL Data}} &
\centering\textbf{Spider Dev} &
\centering\textbf{Spider Test} &
\centering\textbf{BIRD Dev} &
\centering\textbf{Spider DK} &
\centering\textbf{Spider Syn} &
\centering\textbf{Spider Realistic} &
\multirow{2}{\linewidth}{\centering\textbf{Average}} \tabularnewline
\midrule

None & \centering{77.85} & \centering{79.60} & \centering{47.46} & \centering{66.73} & \centering{66.05}  & \centering{70.08} & \centering{67.96} \tabularnewline
\midrule

Spider & \centering{77.27} & \centering{79.74} & \centering{46.74} & \centering{68.04} & \centering{68.67}  & \centering{69.09} & \centering{68.26} \tabularnewline

Spider-Augmented & \centering{80.08} & \centering{82.30} & \centering{49.54} & \centering{71.78} & \centering{71.18}  & \centering\textbf{75.79} & \centering{71.78} \tabularnewline
\midrule

BIRD & \centering{79.30} & \centering{80.48} & \centering{51.62} & \centering{69.35} & \centering{70.31}  & \centering{73.82} & \centering{70.81} \tabularnewline

BIRD-Augmented & \centering{79.59} & \centering{80.67} & \centering{51.56} & \centering{69.35} & \centering{70.79}  & \centering{70.67} & \centering{70.44} \tabularnewline
\midrule

\datasetnametable{} & \centering\textbf{82.69} & \centering\textbf{82.53} & \centering\textbf{53.13} & \centering\textbf{71.96} & \centering\textbf{72.24}  & \centering{73.62} & \centering\textbf{72.70} \tabularnewline

\bottomrule
\end{tabular}
\caption{Comparison of execution accuracy (\%) on Spider, BIRD, and the Spider variants for data augmentation settings in which RingSQL data was mixed in with the Spider and BIRD data. Base model used for training and evaluation in all settings was Qwen2.5-Coder-3B-Instruct.}
\label{tab:augmentation_results}
\end{table*}

\section{Impact of SQR Template Diversity}
\label{app:sqr_diversity}

\begin{table}[ht!]
\centering
\footnotesize
\setlength{\tabcolsep}{5pt}
\begin{tabular}{lcccc}
\toprule
\textbf{RL Data} 
  & \shortstack{\textbf{Spider}\\\textbf{Dev}} 
  & \shortstack{\textbf{Spider}\\\textbf{Test}} 
  & \shortstack{\textbf{BIRD}\\\textbf{Dev}} 
  & \textbf{Avg.} \\
\midrule
\datasetnametable{}  & 82.69 & 82.53 & 53.13 & 72.78 \\
\midrule
Only Generators      & 79.40 & 80.76 & 53.56 & 71.24 \\
Only Templates       & 74.37 & 74.10 & 45.76 & 64.74 \\
\bottomrule
\end{tabular}
\caption{Execution accuracy (\%) on Spider and BIRD when using different template mixtures for data generation. Base model: Qwen2.5-Coder-3B-Instruct.}
\label{tab:template_variation_ablation}
\end{table}

We conduct an ablation study that varies the diversity and composition of the SQR templates and generators used to construct the training data. Specifically, we compared (1) SQR templates only, (2) templated generators only, and (3) \datasetname{} which combines both. Each dataset consists of 4,000 training and 1,000 development examples. Table \ref{tab:template_variation_ablation} provides evaluation results for each setting for the Spider and Bird benchmarks. The generators provide broad coverage across query types and structural patterns, while the SQR templates target more specific and complex query constructions. As expected, using only the targeted templates resulted in weaker performance than the broader generator-based set. However, combining both sources produced the strongest models, suggesting that diversity and structural coverage are important factors in effective training. A more systematic exploration of template diversity and mixing strategies could yield additional gains, and we view this as a promising direction for future work.

\section{Spider 2.0-lite Evaluation}
\label{app:spider_2}

Spider 2.0 \cite{lei2025spider} is a benchmark consisting of text-to-SQL workflow problems derived from enterprise-level database use cases, featuring queries and natural language formulations that are substantially more complex than those found in earlier benchmarks such as Spider and BIRD. Achieving strong performance on this benchmark typically requires large-scale systems built around multi-stage pipelines, encompassing schema linking, filter value linking, candidate SQL generation and evaluation, and iterative refinement often orchestrated with agentic techniques \cite{somayajula2026somasqlresolvingmultisourceambiguity, deng2025reforcetexttosqlagentselfrefinement, wang2025autolinkautonomousschemaexploration}.

Our work takes a different aim: rather than building such end-to-end systems, we focus on strengthening the foundational SQL reasoning ability that underlies them. This is a complementary and necessary contribution, particularly as reinforcement learning becomes an increasingly important stage in modern post-training pipelines. Nevertheless, to provide a complete picture, we evaluate the effect of training with our range of synthetic and human-annotated datasets across multiple model architectures and report accuracy on Spider 2.0-lite. Results are reported in Table \ref{tab:spider2_results}.

Observed accuracies are consistent with those reported in other recent work \cite{ma2025sql}, but remain too low across all configurations to surface any meaningful signal regarding which training datasets or approaches are most beneficial. This is unsurprising: the relatively small model architectures used in our study are not inherently capable enough for this benchmark, nor are they embedded in the multi-stage pipelines that state-of-the-art Spider 2.0 systems depend on. We therefore treat these results as a reference point rather than a basis for drawing conclusions about dataset quality, and leave integration into larger pipeline architectures to future work.

\begin{table}[ht!]
\centering
\footnotesize
\setlength{\tabcolsep}{5pt}
\begin{tabular}{llc}
\toprule
\textbf{Base Model} & \textbf{RL Data} 
  & \shortstack{\textbf{Spider 2.0-lite Sqlite}} \\
\midrule
\multirow{8}{*}{\shortstack[l]{Llama-3.2\\3B-Instruct}}
& None                & 0.74 \\
\cmidrule(lr){2-3}
& SynSQL-Unif         & \textbf{1.48} \\
& SynSQL-Comp         & 0.0 \\
& SENSE               & 0.74 \\
& SQL-Flow            & 0.74 \\
& \datasetnametable{} & \textbf{1.48} \\
\cmidrule(lr){2-3}
& Spider              & \underline{1.48} \\
& BIRD                & \underline{1.48} \\
\midrule

\multirow{8}{*}{\shortstack[l]{Qwen2.5-Coder\\3B-Instruct}}
& None                & 2.22 \\
\cmidrule(lr){2-3}
& SynSQL-Unif         & 3.70  \\
& SynSQL-Comp         & \textbf{5.19} \\
& SENSE               & 3.70  \\
& SQL-Flow            & \textbf{5.19} \\
& \datasetnametable{} & \textbf{5.19} \\
\cmidrule(lr){2-3}
& Spider              & 2.96 \\
& BIRD                & 4.44 \\
\midrule

\multirow{8}{*}{\shortstack[l]{Llama-3.1\\8B-Instruct}}
& None                & 2.96  \\
\cmidrule(lr){2-3}
& SynSQL-Unif         & 2.22  \\
& SynSQL-Comp         & 3.70 \\
& SENSE               & \textbf{4.44} \\
& SQL-Flow            & 3.70 \\
& \datasetnametable{} & 3.70 \\
\cmidrule(lr){2-3}
& Spider              & 2.96 \\
& BIRD                & 2.22 \\
\midrule

\multirow{8}{*}{\shortstack[l]{Qwen2.5-Coder\\7B-Instruct}}
& None                & 8.15 \\
\cmidrule(lr){2-3}
& SynSQL-Unif         & 9.63 \\
& SynSQL-Comp         & \textbf{11.11} \\
& SENSE               & 6.67 \\
& SQL-Flow            & 9.63 \\
& \datasetnametable{} & 8.15 \\
\cmidrule(lr){2-3}
& Spider              & 7.41 \\
& BIRD                & 8.89 \\
\bottomrule
\end{tabular}
\caption{Execution accuracy (\%) on Spider 2.0-lite sqlite test set. Within each model, rows are grouped as: no RL training (top), synthetic RL data (middle), and gold-standard RL data (bottom, upper bound). Best result among synthetic data per model shown in \textbf{bold}. Results for non-synthetic data that perform at least as well as all synthetic data tuned models per model are \underline{underlined}.}
\label{tab:spider2_results}
\end{table}

\end{document}